\documentclass{article}
\PassOptionsToPackage{numbers, compress}{natbib}
\usepackage[preprint]{neurips_2025}

\usepackage{array}
\usepackage[utf8]{inputenc} 
\usepackage[T1]{fontenc}    
\usepackage{hyperref}       % hyperlinks
\usepackage{url}            % simple URL typesetting
\usepackage{booktabs}       % professional-quality tables
\usepackage{amsfonts}       % blackboard math symbols
\usepackage{nicefrac}       % compact symbols for 1/2, etc.
\usepackage{microtype}      % microtypography
\usepackage{xcolor}         % colors
\usepackage{graphicx}       
\usepackage{todonotes}
\usepackage{amsmath}
\usepackage[ruled, linesnumbered]{algorithm2e}
\usepackage{wrapfig}
\usepackage{xspace}
\usepackage{multirow}
\usepackage{cleveref}
\usepackage{setspace}
\usepackage{float}
\usepackage{subfigure}
\usepackage{subcaption}
\newcommand{\ourMethod}{SALE\xspace}
\newcommand{\lowBitPass}{Selection-Pass\xspace}
\newcommand{\fullPass}{Computation-Pass\xspace}
\newcommand{\cwidth}[0]{0.8cm}
\newcommand{\prefill}[0]{prefilling\xspace}

\title{SALE :  Low-bit Estimation for Efficient Sparse Attention in Long-context LLM Prefilling}

\author{%
 Xiaodong Ji\textnormal{,} Hailin Zhang\textnormal{,} Fangcheng Fu\textnormal{,} Bin Cui \\
 Peking University \\
 \texttt{xiaodong.0731@stu.pku.edu.cn\textnormal{,} z.hl@pku.edu.cn\textnormal{,} }\\\texttt{ccchengff@pku.edu.cn\textnormal{,} bin.cui@pku.edu.cn}
}

\begin{document}

\maketitle

\begin{abstract}
Many advanced Large Language Model (LLM) applications require long-context processing, but the self-attention module becomes a bottleneck during the \prefill stage of inference due to its quadratic time complexity with respect to sequence length. 
Existing sparse attention methods accelerate attention computation by skipping less significant regions of the attention map. 
However, these approaches typically perform coarse-grained inspection of the attention map, rendering considerable loss in model accuracy.
In this paper, we propose \textbf{\ourMethod}, a fine-grained sparse attention method that accelerates the long-context \prefill stage of LLM with negligible loss in model accuracy. 
\ourMethod achieves fast and accurate fine-grained attention weight estimation through 4-bit quantized query-key products, followed by block-sparse attention to accelerate \prefill computations. 
For importance evaluation for query-key pairs, we adopt our \textit{Relative Attention Score} metric, which offers significantly higher efficiency within our framework.
We implement a custom CUDA kernel optimized for our approach for hardware efficiency, reducing the additional overhead to approximately 11$\%$ of the full attention latency.
Notably, \ourMethod requires no parameter training and can be seamlessly integrated into existing systems with trivial code modifications. 
Experiments on long-context benchmarks demonstrate that our method outperforms existing approaches in accuracy-efficiency trade-offs,
achieving at least 3.36$\times$ speedups on Llama-3.1-8B for sequences longer than 64K while maintaining model quality.
Our code is available at~\href{https://github.com/BirdChristopher/SALE}{\texttt{https://github.com/BirdChristopher/SALE}}.
\end{abstract}

\section{Introduction}
Due to the demand for ultra-long context understanding in many complex applications such as long book summarization~\cite{kryściński2022booksumcollectiondatasetslongform, porwal2023transformer, chang2024booookscore},  long document question-answering~\cite{caciularu-etal-2023-peek, pang2022qualityquestionansweringlong, fan2019eli5longformquestion}, and repository-level code completion~\cite{wang2024teachingcodellmsuse, wang2024rlcoderreinforcementlearningrepositorylevel}, state-of-the-art Large Language Models~(LLM) are now capable of supporting increasingly longer context window~\cite{grattafiori2024llama3herdmodels, yang2025qwen2, gemmateam2025gemma3technicalreport, deepseekai2025deepseekv3technicalreport}.
Most of LLMs adopt decoder-only Transformer architecture~\cite{vaswani2017attention}, in which the self-attention module serves as the core component that enables powerful language understanding capabilities.
However, during the prefill stage of LLM inference, the self-attention module exhibits quadratic time complexity with respect to the input token count, causing computational costs to rise rapidly for long texts and becoming the primary bottleneck~\cite{fu2024challengesdeployinglongcontexttransformers, jiang2024minference}.

In recent years, numerous research studies have attempted to accelerate attention by computing only the important regions of attention maps, based on the observation that attention maps in LLMs are significantly sparse~\cite{DBLP:journals/corr/abs-2404-02690}. 
These methods are referred to as \textit{sparse attention}, and they use \textit{sparse masks} to indicate the regions of attention map that are computed.
Some sparse attention methods leverage sparse masks with static patterns, including stride pattern~\cite{child2019generating}, window pattern~\cite{ zaheer2020big, beltagy2020longformer}, and streaming pattern~\cite{ xiao2024efficient, han2023lm}.
However, static sparse masks often result in severe performance degradation, as the sparse patterns within LLM attention maps are highly dynamic across various input content.
To construct sparse masks that can dynamically adapt to different input, some approaches, such as MInference~\cite{jiang2024minference} and SampleAttention~\cite{zhu2024sampleattention}, decompose the sparse attention pattern into combinations of multiple vertical or slash lines, and predict the positions of these lines by analyzing the attention score distribution of a subset of query tokens.
Another series of sparse attention methods, such as FlexPrefill~\cite{lai2025flexprefill}, SpargeAttn~\cite{zhang2025spargeattn}, and HiP Attention~\cite{lee2025a}, view the attention map as the concatenation of multiple blocks and dynamically skip certain attention computation at the block granularity.
These methods often construct a representative token for each consecutive query/key token chunk and build sparse masks based on the product between representative tokens.
Although existing dynamic sparse attention methods can accelerate the prefilling stage of LLM to some extent, they fail to achieve an satisfactory accuracy-efficiency trade-off since they cannot perform fine-grained inspection of the entire attention map.

In this paper, we propose \textbf{\ourMethod}, a novel training-free block-\textbf{S}parse \textbf{A}ttention technique based on \textbf{L}ow-bit \textbf{E}stimation of attention weights, to significantly accelerate the long-context prefilling stage of LLM with negligible loss in model accuracy.
Unlike existing approaches that rely on coarse-grained approximations of the attention map, \ourMethod performs fine-grained element-wise analysis, enabling the construction of highly sparse attention masks while bounding output error within an acceptable tolerance.
To minimize the additional overhead introduced by fine-grained inspection, \ourMethod uses 4-bit quantized query and key vectors to approximate attention weights. 
This computation can be executed efficiently on modern GPUs due to the use of high-throughput low-bit Tensor Core instructions and reduced global memory access. 
Furthermore, motivated by the observation that the attention weights in the \textit{sink} (beginning) and \textit{local} (end) regions of each attention map row tend to be relatively higher~\cite{xiao2024efficient, gu2025when}, we design a novel metric \textit{Relative Attention Score} which reflects the relative magnitude between the current attention weight and those within the sink-local region. 
Compared to common practice that uses attention scores ~\cite{zhang2023h2o, li2024snapkv, zhang2024pqcache, liu2025speculative} as the indicator for pruning, our approach introduces negligible computational overhead and is able to adaptively adjust the sparsity level based on the input.
In addition to our algorithm design, we introduce several kernel optimization techniques that further improve the hardware efficiency. 
The results of the speed test show that our custom CUDA implementation of attention map inspection takes only 11$\%$ of the computation time of full attention.

We conduct comprehensive experiments on various long-context processing benchmarks using Llama-3.1-8B-Instruct~\cite{grattafiori2024llama3herdmodels} and Qwen-2.5-32B-Instruct~\cite{yang2024qwen2} to verify the effectiveness of our method. Experimental results demonstrate that our method delivers a speedup of at least 3.36$\times$ when processing sequences longer than 64K tokens, while maintaining negligible accuracy loss. It achieves superior accuracy-efficiency trade-off compared to baseline methods.

\section{Related works} 
\label{sec:related_works}
\paragraph{Sparse LLM prefilling}
Many previous works try to leverage the sparsity nature of transformer model to accelerate LLM inference from different perspectives.

One line of research exploits input text sparsity to dynamically prune context irrelevant to the user's query~\cite{jha2024characterizing,liu2025speculative, shi2024discovering, jiang2023llmlingua, li2023compressing}.
While these methods can significantly reduce LLM inference latency for relatively simple prompts,  they severely degrade generation quality when processing complex inputs~\cite{yuan2024kv}.
 
Numerous studies have observed sparsity patterns in self-attention modules, where only a small subset of attention map elements are much larger than the rest.
Some methods~\cite{xiao2024efficient, fu2024moa, xiao2024duoattention} use predefined static sparsity patterns to prune the attention map. 
However, these methods suffer from accuracy degradation as the attention sparsity distribution varies among different input contexts~\cite{jiang2024minference,lai2025flexprefill}. 
Other methods assume that the distribution follows certain structures, such as Vertical-Slash or Block-Sparse. 
Some of them~\cite{jiang2024minference, zhu2024sampleattention} try to dynamically predict the location of important regions by examining the exact attention scores of several tokens. 
Others~\cite{gao2024seerattention, zhang2025spargeattn, lai2025flexprefill, lee2025a} regard the attention map of compressed tokens, which are generated from continuous token chunks, as the proxy of real attention map.
All these methods fail to achieve accurate predictions due to their overly coarse-grained approximations of attention maps.

In contrast to the aforementioned approaches, several alternatives to self-attention have emerged to circumvent its quadratic complexity. Notable examples include: (1) natively sparse attention algorithms~\cite{yuan2025native,lu2025moba}, (2) linear attention mechanisms~\cite{peng2023rwkv,yang2023gated}, and (3) state-space models~\cite{gu2024mamba,10.5555/3692070.3692469}. However, these methods impose significant adoption costs as they necessitate full model retraining.

During the decoding stage, methods like SparQ~\cite{ribar2023sparq} and InfiniGen~\cite{lee2024infinigen} compress the channels of query / key tokens to efficiently approximate the attention scores. 
Retrieval-based approaches~\cite{zhang2024pqcache, chen2025magicpig, liu2024retrievalattention} leverage vector-retrieval technique to approximately sort the attention scores of input tokens. 
Several existing algorithms compress tokens by analyzing attention maps during the prefilling stage. These approaches either eliminate redundant tokens~\cite{zhang2023h2o,liu2023scissorhands,li2024snapkv,ge2023model,devoto2024simple} or perform token merging~\cite{zhang2024cam,zandieh2024subgen}. Our method is orthogonal to these optimizations and can be combined to further enhance end-to-end LLM inference efficiency.

\paragraph{Attention kernel optimization}
Many CUDA kernel optimization techniques~\cite{dao2022flashattention, dao2023flashattention, shah2024flashattention, sanovar2024lean} leverage hardware features to accelerate the computation of the original full attention. Although these methods accelerate computation, they still require full attention calculations and fail to fully exploit the inherent sparsity of attention maps. 
\section{Method}
\label{sec:methodology}

This section presents the detailed architecture of \ourMethod, which operates through three sequential processing stages: \textbf{Quantization}, \textbf{\lowBitPass} and \textbf{\fullPass}.
During \lowBitPass, we select important attention regions at the block granularity and record the coordinates of these blocks. 
We then compute attention on selected blocks in the following \fullPass.

\subsection{Problem formulation}
\label{sec:formulation}
We denote the query, key and value matrix as $Q$, $K$ and $V$, respectively, while the corresponding token at the offset $i$ are $q_i, k_i, v_i$.  Let $N$ represents the sequence length, and $d$ represents the hidden size. 
The shapes of $Q$, $K$ and $V$ are all $N\times d$. 
Single-head self-attention module can be mathematically formalized as below:
\begin{equation}
    Attn(Q, K, V, M) = Softmax(\frac{QK^T}{\sqrt{d}} + M) V
\end{equation}
During the computation of self-attention, attention weight matrix $S$ is defined as $S = {QK^T}/{\sqrt{d}}$, and attention score matrix $P$ is defined as $P = Softmax(S + M)$.
Matrix $M$ is the sparse attention mask with a shape of $N\times N$. 
It is formed by $M = M_c + M_s$, where $M_c, M_s \in \{0,-\infty \}$ represent the causal mask and sparse mask respectively. 
Based on the mathematical properties of Softmax function, if an item $M[i,j]$ in matrix $M$ is $-\infty$, its corresponding attention score will be zero. Therefore, we can skip the attention computation at this position.

For block-sparse attention, query and key tokens are divided into continuous blocks of sizes $b_q, b_k$ along the sequence length dimension. We denote the query, key token block at position $j$ as $Q_j$, $K_j$, which have shapes of $b_q \times d$ and $b_k \times d$ respectively. 
For simplicity, we assume $b_q \mid N$, $b_k \mid N$, and denote $N_q = N/b_q$, $N_k = N/b_k$. As shown in~\Cref{fig:demo:workflow}, the attention map can be viewed as the concatenation of $N_q\cdot N_k$ attention blocks, each of shape $b_q \times b_k$. 
Block sparse attention skips computation at the block level. To formulate, we denote $M_{bs} \in \{0,1\}$ as \textit{block-level sparse mask}, and values of sparse mask $M_s$ depend on $M_{bs}$:
\begin{equation}
M_s[i,j] = 
    \begin{cases} 
        0, & \text{if } \quad M_{bs}[\ \lfloor i / b_q \rfloor,\lfloor j / b_k \rfloor \ ] = 1, \\[3pt]
        -\infty, & \text{if } \quad M_{bs}[\ \lfloor i / b_q \rfloor,\lfloor j / b_k \rfloor \ ] = 0
    \end{cases}
\end{equation}
In other words, the attention computation between $Q_i, K_j, V_j$ will be skipped if $M_{bs}[i,j]$ is zero. 
Block-sparse attention aims to maximize sparsity in matrix $M_{bs}$ while bounding the approximation error relative to full attention within a tolerable threshold.

\subsection{Block selection via fine-grained importance approximation}
\label{sec:algo_design}
We construct $M_{bs}$ during \lowBitPass, which is illustrated in~\Cref{alg:selection_pass}. In order to achieve the optimization objectives for $M_{bs}$ while minimizing additional overhead, \ourMethod proposes two key techniques: \textit{4-bit Attention Weight Approximation} and \textit{Relative Importance Approximation}. 

\paragraph{4-bit attention weight approximation}
To obtain a finer-grained estimation of the attention map, \ourMethod examines the attention weights for all positions. 
The overall computation process of \lowBitPass is akin to that of FlashAttention2~\cite{dao2023flashattention}. Specifically, in the outer loop, we iterate through all query blocks, while in the inner loop, we examine the attention weights between each query block and key block.

During block-wise inspection, rather than full-precision floating-point $Q$ and $K$ matrices, \ourMethod computes attention weights using 4-bit quantized versions $\widetilde{Q}$ and $\widetilde{K}$ to make the approximation.
This design significantly minimizes additional overhead with high-throughput low-bit Tensor Core instructions and reduced GPU global memory access.
In addition, the quantization overhead is negligible. In our implementation, we leverage the quantization algorithm proposed by SageAttention-2~\cite{zhang2024sageattention2}.

\begin{figure}[t]
\vspace{-10pt}
\centering
\subfigure[Attention maps.]{
\scalebox{0.35}{
\label{fig:demo:map}
\includegraphics[width=\linewidth]{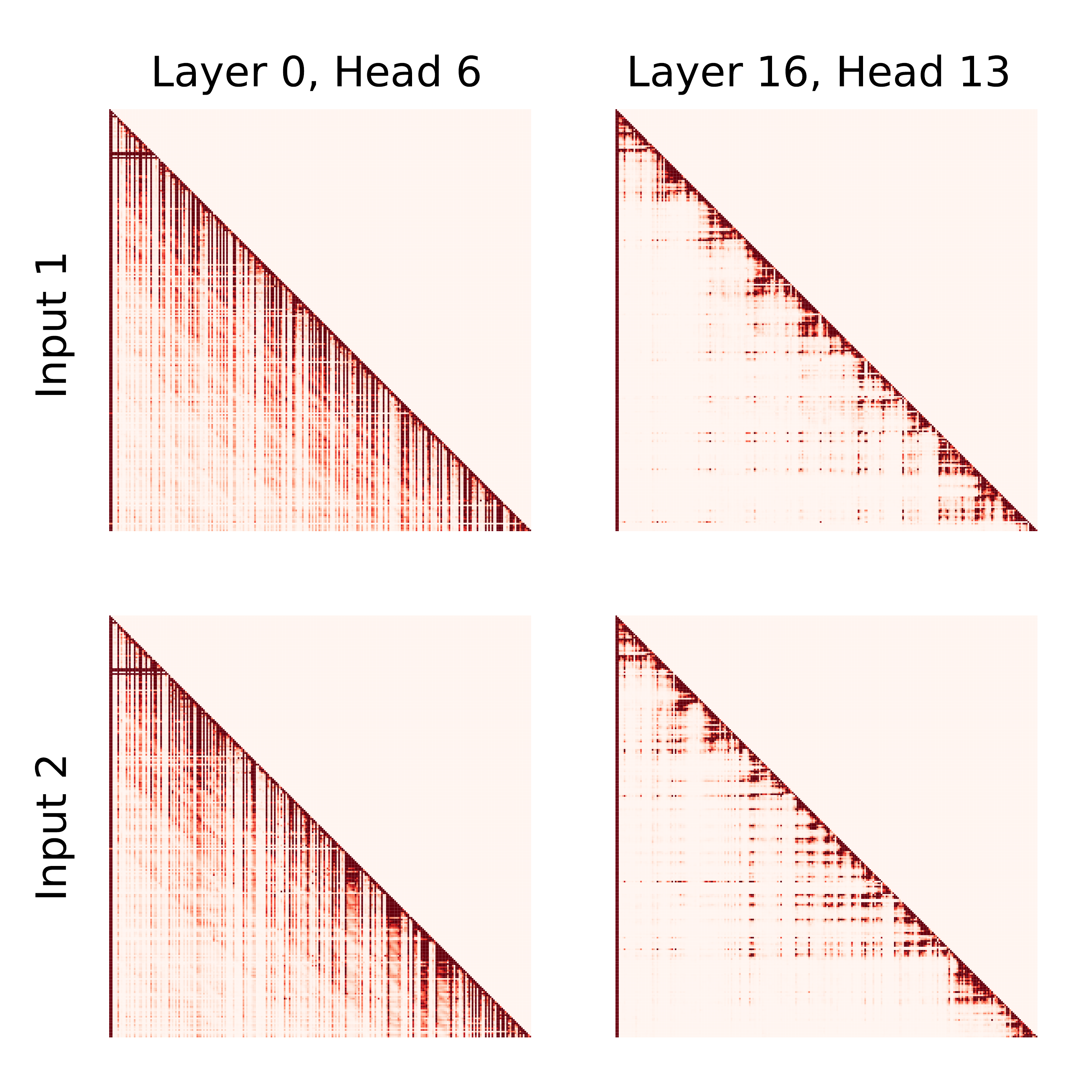}
}
}
\subfigure[Sparse mask obtained after estimation.]{
\scalebox{0.6}{
\label{fig:demo:workflow}
\includegraphics[width=\linewidth]{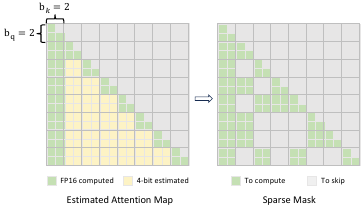}
}
}
\caption{(a) Attention maps of two different attention heads in Llama-3.1-8B-Instruct when processing different input sequences. (b) Illustration of \ourMethod. The whole $16\times16$ attention map is viewed as concatenation of many $2\times2$ blocks. We first estimate the attention weights in an element-wise manner, and then construct a sparse mask at the block level based on these estimations. }
\label{fig:demo}
\vspace{-10pt}
\end{figure}

\paragraph{Relative importance estimation}
Denoting approximated attention weights as $\widetilde{S}$, the next step is to evaluate the ``importance'' of each attention block.
In related works~\cite{zhang2023h2o, li2024snapkv, liu2025speculative, liu2023scissorhands}, a commonly used metric is the attention score, obtained by applying \textit{Softmax} function to attention weights. To perform sparse attention computation which cannot obtain full attention scores, we propose \textit{Relative Attention Score} as our importance metric.
Our design is based on an observation in many related studies~\cite{xiao2024efficient, gu2025when, xiao2024duoattention}. 
As shown in \Cref{fig:demo:map}, attention scores within the ``sink-local'' region~(i.e. the beginning and end of each row) maintain consistently high values, while the region exhibits consistent size across diverse input sequences.
Motivated by this pattern, we assess ``importance'' by comparing $\widetilde{S}[i,j]$ with the attention weights within the sink-local region.
As illustrated in ~\Cref{alg:selection_pass}, before examining blocks located in the middle of the sequence, we first compute full precision attention on blocks in the sink-local area. Denoting the indices of key tokens within the sink-local region as $I_{SL}$, this process yields two intermediate values, $\widetilde{m}_i$ and $\widetilde{l}_i$, which can be formulated as follows:
\[
\widetilde{m}_i = \max_{j \in I_{SL}} S[i,j], \quad \widetilde{l}_i = \sum_{j \in I_{SL}} e^{S[i,j] - \widetilde{m}_i}
\]
Then, the \textit{Relative Attention Score} $\widetilde{P}[i,j]$ can be computed as:
\begin{equation}
  \widetilde{P}[i,j] = \frac{e^{\widetilde{S}[i,j] - \widetilde{m}_i}}{\widetilde{l}_i}
\end{equation}
If all $\widetilde{P}[i,j]$ values in a block are smaller than the threshold $\tau$~(e.g. 0.004), this block is marked as non-critical, and the computation for this block will be skipped in the subsequent \fullPass. The procedure for determining the threshold value $\tau \in (0,1)$ is elaborated in ~\Cref{sec:calib}.

\begin{algorithm}[!ht]
\footnotesize
\setstretch{1.05}
  \caption{\lowBitPass}\label{alg:selection_pass}
  \KwIn{$Q, K \in \mathbb{R}^{N\times d}$ , 4-bit quantized matrices $\widetilde{Q}, \widetilde{K} \in \mathbb{Z}^{N\times d}$, threshold $\tau$, block size $b_q, b_k$, local area size $l$.}
  $N_q \leftarrow N/b_q$, $N_k \leftarrow N/b_k$, $N_{local} \leftarrow l / b_{k}$\;
  Split $Q, K$ into blocks $Q_i \in\mathbb{R}^{b_q\times d}$, $K_j\in\mathbb{R}^{b_k\times d}$, split $ \widetilde{Q},  \widetilde{K}$ into blocks $\widetilde{Q}_i \in\mathbb{Z}^{b_q\times d}$, $\widetilde{K}_j\in\mathbb{Z}^{b_k\times d}$
  \For{$ i = 0$ to $N_q - 1$}{
    $I_{SL} \leftarrow \{0\} \cup [i - N_{local}, i-1]$ \tcp*[r]{Block indices of sink-local area} \
    $\widetilde{m}, \widetilde{l} \in \mathbb{R}^{b_q}$, $\widetilde{m} \leftarrow -\infty$, $\widetilde{l} \leftarrow 0$ \tcp*[r]{Initialize intermediate result} \
    \For{$j \in I_{SL}$}{
        \If{$j\ne0$}{
            $\widetilde{m}_{\Delta} \leftarrow$ $\widetilde{m}$ \space - \space rowmax($Q_iK_j^{T}$ / $\sqrt{d}$); 
            \quad $\widetilde{l} \leftarrow \widetilde{l}$ $\cdot $ exp $(\widetilde{m}_{\Delta}) $ \;
        }
        $\widetilde{m} \leftarrow $ rowmax($Q_iK_j^{T}$ / $\sqrt{d}$) \tcp*[r]{Ignore causal mask} \
        $\widetilde{l} \leftarrow \widetilde{l}$ + rowsum(exp($\frac{Q_iK_j^{T}}{\sqrt{d}}$ - $\widetilde{m}$)) \;
        $M_{bs}[i,j] \leftarrow 1$
    }
    \For{$j \leftarrow 1$ to $(i - N_{local}-1)$}{
        $\widetilde{S}_{ij} \leftarrow$ Dequantize($\widetilde{Q}_i\widetilde{K}_j^{T}$) $/ \sqrt{d}$ \tcp*[r]{Approximate attention weight} \
        $\widetilde{P}_{ij} \leftarrow$ exp$(\widetilde{S}_{ij} - \widetilde{m})$ / $\widetilde{l} $ \tcp*[r]{Compute Relative Attention Score} \
        $M_{bs}[i,j] \leftarrow $ max$(\widetilde{P}_{ij}) \ge \tau $ \;
    }
  }
  \KwOut{Block-level sparse mask $M_{bs}$}
\end{algorithm}

\subsection{Per-head threshold calibration}
\label{sec:calib}
\Cref{fig:demo:map} illustrates the attention score distributions of two attention heads of Llama-3.1-8B-Instruct, exhibiting inconsistent sparsity levels.
Thus, applying the same $\tau$ for all heads may lead to suboptimal performance. To address the issue, we propose an offline calibration procedure to determine the optimal $\tau$ value for each head, which ensures negligible output errors while maximizing sparsity.

We adopt the $L_1$ distance between the output of \ourMethod and the output of full attention as the error metric, which can be formulated as $Err(\tau) = \|O-\widetilde{O}\|_1 / N$ . $O$ is the result of the original attention, $\widetilde{O}$ is the result of \ourMethod, and $N$ represents sequence length.
At the beginning of the calibration, $\tau$ is initially set to be a relatively large threshold $\tau_{0}$~(e.g. 0.008). We then progressively reduce the sparsity level by halving the value of $\tau$ until $Err(\tau)$ falls below $\theta$, where $\theta$ is the predefined error bound. By tuning $\theta$, we can control the sparsity level of \ourMethod.

\subsection{Kernel optimization}
\paragraph{Reduction in dequantization operations} 
Theoretically, whether an attention block is skipped only depends on the comparison between the largest \textit{Relative Attention Score} with $\tau$. By employing per-thread quantization strategy proposed in ~\cite{zhang2024sageattention2}, we make all quantized attention weight elements held by each thread share the same quantization scale. This ensures that the largest \textit{Relative Attention Score} and the largest approximated attention weight occur at the same position. 
Therefore, only the largest approximated attention weight needs to be dequantized, which saves many low-throughput operations such as datatype conversion.

\paragraph{Relative attention score comparison}
Directly computing \textit{Relative Attention Score} is time-consuming as it consists of multiple complex hardware instructions, including floating point division and exponential function. Considering that $\widetilde{l}_i$ and $\widetilde{m}_i$ do not change after the computation in sink-local area, we optimize this comparison by following mathematical transformation:
\begin{equation}
  \frac{e^{\widetilde{S}[i,j] - \widetilde{m}_i}}{\widetilde{l}_i} \ge \tau \iff \widetilde{S}[i,j] \ge \ln(\tau \cdot \widetilde{l}_i +  \widetilde{m}_i)
\end{equation}
The comparison between the \textit{Relative Attention Score} and $\tau$ can then be accomplished using a single floating point comparison instruction. 
It is worth noting that we also mitigate potential overflow issues caused by the exponential function.

\paragraph{Integration with SageAttention}
\label{sec:sage_up}
The final stage of \ourMethod is \fullPass. In this stage, sparse attention is computed only on the important blocks selected by \lowBitPass.
We employ the QKV quantization strategy proposed in SageAttention~\cite{zhang2025sageattention} to further accelerate \fullPass while maintaining negligible precision loss.

\section{Experiments}
\label{sec:expr}
\subsection{Settings}
\paragraph{Models}
Most of the experiments are conducted using Llama-3.1-8B-Instruct~\cite{grattafiori2024llama3herdmodels}~(\textbf{Llama-3.1}). We also use Qwen2.5-32B-Instruct~\cite{yang2024qwen2}~(\textbf{Qwen-2.5}) to validate the effectiveness of our method on larger-scale LLM. Both of these models support context lengths of 128K. We use the default chat template to construct the input prompt.

\paragraph{Implementation details}
We implement \lowBitPass in C++ CUDA and use Triton~\cite{10.1145/3315508.3329973} compiler to accelerate the quantization process. We implement the quantized \fullPass based on the open-source code of SpargeAttn~\cite{zhang2025spargeattn}. For model inference, we leverage the transformers~\cite{wolf-etal-2020-transformers} library to build an execution pipeline and replace the default self-attention module with \ourMethod. We use greedy decoding to avoid randomness during generation.
For those hyper-parameters mentioned in~\Cref{sec:algo_design}, we use block size $b_q = 64$ and $b_k = 32$. For the sink-local area discussed in~\Cref{sec:algo_design}, we constrain the sink area size to 32 tokens and the local area size to no more than 256 tokens for any input sequence. During offline calibration, we set the initial threshold $\tau_0 = 0.008$, and use error bounds of $\theta = 0.4$ for Llama-3.1 and $\theta=2.0$ for Qwen-2.5 by default. 
All latency experiments are conducted on a server with 8 GeForce RTX 4090 GPUs without using tensor-parallel~\cite{shoeybi2019megatron} or context-parallel~\cite{li2023distflashattn} technique.

\paragraph{Baselines}
\label{sec:baselines}
To demonstrate the advantages of \ourMethod, we compare it with four strong baselines for self-attention acceleration in long-context processing: \textbf{FlashAttention2}(\textit{FA2})~\cite{dao2023flashattention}, \textbf{MInference}(\textit{MInfer})~\cite{jiang2024minference}, \textbf{FlexPrefill}(\textit{Flex})~\cite{lai2025flexprefill}, and \textbf{SpargeAttn}(\textit{Sparge})~\cite{zhang2025spargeattn}.
FA2 computes standard full attention, while the other three methods employ sparse attention mechanisms.
All experimental results are based on their publicly available implementation. We use $\gamma = 0.95$ for both Llama-3.1 and Qwen-2.5 when evaluating FlexPrefill. We use $(l_1 = 0.08,l_2=0.09)$ for Llama-3.1, and $(l_1 = 0.04, l_2 = 0.05)$ for Qwen-2.5 when evaluating SpargeAttn. For MInference, we select the sparse pattern for each head based on its open-source code.

Additionally, to investigate the performance of these methods under varying sparsity levels, we prepare multiple sets of hyperparameters based on their publicly available codes.
For FlexPrefill and SpargeAttn, as described in their papers, we adjust their sparsity levels by tuning $\gamma$ and $(l_1, l_2)$,  respectively. 
For MInference, since its open-source implementation configures all heads with the \textit{Vertical-Slash} pattern, the sparsity rate is adjusted by varying the total number of vertical and slash lines across all heads.
To ensure a fair comparison, we use the calibration samples for MInference, SpargeAttn, and \ourMethod.

\paragraph{Metrics}
To validate the effectiveness of \ourMethod, we assess model quality using long-context benchmarks (see~\Cref{sec:expr:acc}) and quantify efficiency through latency measurements.
All latency results in the experimental section focus solely on the attention computation time across \underline{all layers} during the LLM \prefill phase. Our latency measurements include all online operations, such as quantization, block selection, and index selection. In some experiments, we report the end-to-end~(E2E) latency on certain datasets, which is computed by summing the latency of all samples in the dataset.

\subsection{Accuracy evaluation}
\label{sec:expr:acc}
Following common practice~\cite{zhang2025spargeattn, jiang2024minference, lai2025flexprefill, zhang2024pqcache, gao2024seerattention, li2024snapkv}, we adopt three long-context understanding benchmarks to compare the generation quality of our method with other baselines.
These benchmarks employ task-specific evaluation metrics, including accuracy, F1-score, and Rouge-L, where higher values indicate better performance.
(1) \textbf{LongBench}~\cite{bai-etal-2024-longbench}: A comprehensive benchmark covering diverse long-text applications, including single-document QA, multi-document QA, summarization, few-shot learning, synthetic tasks, etc. The context lengths of most input samples are below 32K tokens. 
(2) \textbf{InfiniteBench}~\cite{zhang2024inftybenchextendinglongcontext}: A benchmark designed to evaluate the capability of processing excessively long context (exceeding 100K tokens). It comprises several challenging synthetic tasks such as Retrieve.KV and Math.Find, as well as other real-world tasks including QA and summarization based on fake books or fake dialogues.
(3) \textbf{Needle-In-A-Haystack}~\cite{Greg2024niah}: A widely-used long-context retrieval task. 
It requires the LLM to locate a randomly inserted sentence at various positions within a real-world context.
For all these benchmarks and tasks, we employ the official evaluation scripts from their respective open-source repositories to assess model outputs.

\paragraph{LongBench}
~\Cref{tab:longbench} presents the LongBench evaluation results comparing \ourMethod with baseline approaches.
In the second row of the table, we use abbreviations introduced in ~\Cref{sec:baselines} to denote each method. 
In the last two rows, we report the average scores as well as the latency speedup achieved when processing 64K-length Needle-In-A-Haystack input.

The results on both models show that our approach not only achieves superior accuracy but also delivers the highest acceleration ratio among all sparse attention baselines. In addition, when applying our method, Llama-3.1 exhibits only marginal performance degradation while Qwen-2.5 shows improvement. 
We attribute this improvement to our method's ability to potentially filter noisy information during the \prefill phase, thereby enhancing the model's comprehension capabilities.

\begin{table}[t]
\scriptsize
\centering
\setlength{\abovecaptionskip}{8pt}
\setlength{\tabcolsep}{5pt}
\setlength{\extrarowheight}{1pt}
\caption{LongBench evaluation results of different methods. We use boldface to denote the highest value and underline to indicate the second-highest value.}
\label{tab:longbench}
\begin{tabular}{|p{1.5cm}<{\centering}|p{\cwidth}<{\centering}p{\cwidth}<{\centering}p{\cwidth}<{\centering}p{\cwidth}<{\centering}p{\cwidth}<{\centering}|p{\cwidth}<{\centering}p{\cwidth}<{\centering}p{\cwidth}<{\centering}p{\cwidth}<{\centering}p{\cwidth}<{\centering}|}
\toprule
 &  \multicolumn{5}{c|}{\textbf{Llama-3.1}} & \multicolumn{5}{c|}{\textbf{Qwen-2.5}} \\
\multirow{-2}{*}{\textbf{Tasks}} & \textbf{FA2} & \textbf{MInfer} & \textbf{Flex} & \textbf{Sparge} & \textbf{\ourMethod} & \textbf{FA2} & \textbf{MInfer} & \textbf{Flex} & \textbf{Sparge} & \textbf{\ourMethod} \\
\midrule 
NarrativeQA     &\textbf{29.93}  &24.92  &28.29  &\underline{29.62}  &28.95  &29.20   &\underline{31.27}   &29.80   &29.19  &\textbf{32.21} \\
Qasper          &\underline{44.82}   &44.29  &44.55  &43.73  &\textbf{45.33}  &\underline{45.79}   &45.05   &45.53   &44.61  &\textbf{45.95} \\
MultiFieldQA    &54.65   &53.71  &\underline{55.34}  &\textbf{56.02} &55.18  &\underline{53.25}   &53.01   &52.61   &51.66  &\textbf{53.37} \\
HotpotQA        &\underline{55.81}   &52.00  &55.38  &54.57  &\textbf{55.83}  &\underline{64.68}   &64.59   &\textbf{64.78}   &63.94  &63.95 \\
2WikiMQA        &\underline{46.16}   &44.10  &43.43  &\textbf{47.08}  &42.61  &60.87   &60.82   &\textbf{62.98}   &61.13  &\underline{62.33} \\
MuSiQue         &\underline{30.41}   &25.72  &30.07  &\textbf{31.40}  &30.10  &39.89   &\textbf{41.38}   &39.46   &39.22  &\underline{40.54} \\
GovReport       &\underline{35.29}   &35.09  &34.64  &35.04  &\textbf{35.45}  &30.38   &30.59   &\textbf{30.78}   &30.36  &\underline{30.66} \\
QMSum           &25.25   &\underline{25.47}  &\textbf{25.83}  &25.12  &25.33  &23.06   &23.16   &23.10   &\underline{23.18}  &\textbf{23.42} \\
TREC            &\textbf{72.50}   &\underline{72.00}  &70.50  &71.00  &70.50  &\underline{73.50}   &\underline{73.50}   &\underline{73.50}   &\textbf{74.50}  &73.00 \\
TriviaQA        &\underline{91.65}   &91.18  &89.81  &\textbf{92.68}  &90.47  &87.68   &88.40   &\textbf{89.40}   &\underline{88.81}  &87.97 \\
SAMSum          &43.67   &\underline{43.73}  &43.18  &43.18  &\textbf{44.19}  &45.67   &45.92   &\textbf{46.43}   &\underline{46.41}  &45.92 \\
LSHT            &\textbf{46.50}   &\underline{46.00}  &41.00  &45.50  &\textbf{46.50}  &45.79   &\textbf{47.50}   &44.17   &47.00  &\underline{47.21} \\
Count           &\underline{6.72}    &3.25   &2.59   &5.89   &\textbf{7.09}   &12.67   &\textbf{13.67}   &3.57    &9.22   &\underline{13.38} \\
Retrieval       &\underline{99.50}   &97.00  &82.00  &84.00  &\textbf{100.00} &\textbf{99.50}   &\underline{99.25}   &92.25   &98.83  &98.25 \\
\midrule
Average         &\textbf{48.77}   &47.03  &46.18  &47.48  &\underline{48.39}  &50.85   &\underline{51.29}   &49.88   &50.57  &\textbf{51.30} \\
\midrule
Speedup (64K)   &1.00$\times$   &1.07$\times$  &2.21$\times$  &\underline{3.11$\times$}  &\textbf{3.36$\times$}  &1.00$\times$   &1.25$\times$   &1.39$\times$   &\underline{2.55$\times$}  &\textbf{3.28$\times$} \\
\bottomrule
\end{tabular}
\vspace{-10pt}
\end{table}
\begin{table}[b]
\scriptsize
\centering
\vspace{-15pt}
\setlength{\abovecaptionskip}{8pt}
\setlength{\tabcolsep}{5pt}
\setlength{\extrarowheight}{1pt}
\caption{InfiniteBench evaluation results of different methods. We use boldface to denote the highest value and underline to indicate the second-highest value.}
\label{tab:infinitebench}

\begin{tabular}{|p{1.5cm}<{\centering}|p{\cwidth}<{\centering}p{\cwidth}<{\centering}p{\cwidth}<{\centering}p{\cwidth}<{\centering}p{\cwidth}<{\centering}|p{\cwidth}<{\centering}p{\cwidth}<{\centering}p{\cwidth}<{\centering}p{\cwidth}<{\centering}p{\cwidth}<{\centering}|}
\toprule
 &  \multicolumn{5}{c|}{\textbf{Llama-3.1}} & \multicolumn{5}{c|}{\textbf{Qwen-2.5}} \\
\multirow{-2}{*}{\textbf{Tasks}} & \textbf{FA2} & \textbf{MInfer} & \textbf{Flex} & \textbf{Sparge} & \textbf{\ourMethod} & \textbf{FA2} & \textbf{MInfer} & \textbf{Flex} & \textbf{Sparge} & \textbf{\ourMethod} \\
\midrule 
Retrieve.KV     &\underline{55.60}  &20.00  &38.00   &47.20   &\textbf{56.40}  &4.00   &\textbf{7.60}   &4.80   &4.60   &\underline{5.40} \\
En.MC           &\underline{67.25}  &55.02  &\textbf{68.56}   &66.38   &66.38  &\textbf{63.70}  &\underline{63.32}  &59.80  &61.50  &62.80 \\
Math.Find       &34.29  &\textbf{34.86}  &30.00   &\underline{34.57}   &30.57  &41.40  &45.71  &\underline{47.20}  &41.70  &\textbf{52.00} \\
En.QA           &\textbf{15.12}  &13.96  &\underline{14.18}   &13.51   &13.19  &6.70   &\underline{6.85}   &6.70   &6.80   &\textbf{6.90} \\
En.Dia          &16.50  &13.50  &17.00   &\underline{17.50}   &\textbf{19.00}  &27.50  &\textbf{30.00}  &\underline{29.00}  &25.50  &27.50 \\
\midrule
Average         &\textbf{37.75}  &27.47 &33.55   &35.83  &\underline{37.11}  &28.66   &\underline{30.70}   &29.5  &28.02  &\textbf{30.92} \\
\midrule
Speedup (64K)   &1.00$\times$  &1.07$\times$  &2.21$\times$   &\underline{3.11$\times$}  &\textbf{3.36$\times$}  &1.00$\times$   &1.25$\times$   &1.39$\times$ &\underline{2.55$\times$}  &\textbf{3.28$\times$} \\
\bottomrule
\end{tabular}
\end{table}

\paragraph{InfiniteBench}
~\Cref{tab:infinitebench} presents the test scores of InfiniteBench, evaluating the capability of processing extremely long inputs. 
As shown in the table, our method also achieves the best accuracy-efficiency trade-off on InfiniteBench.

\begin{figure}[t]
\vspace{0pt}
\centering
\subfigure[FlashAttention2.]{
\scalebox{0.3}{
\includegraphics[width=\linewidth]{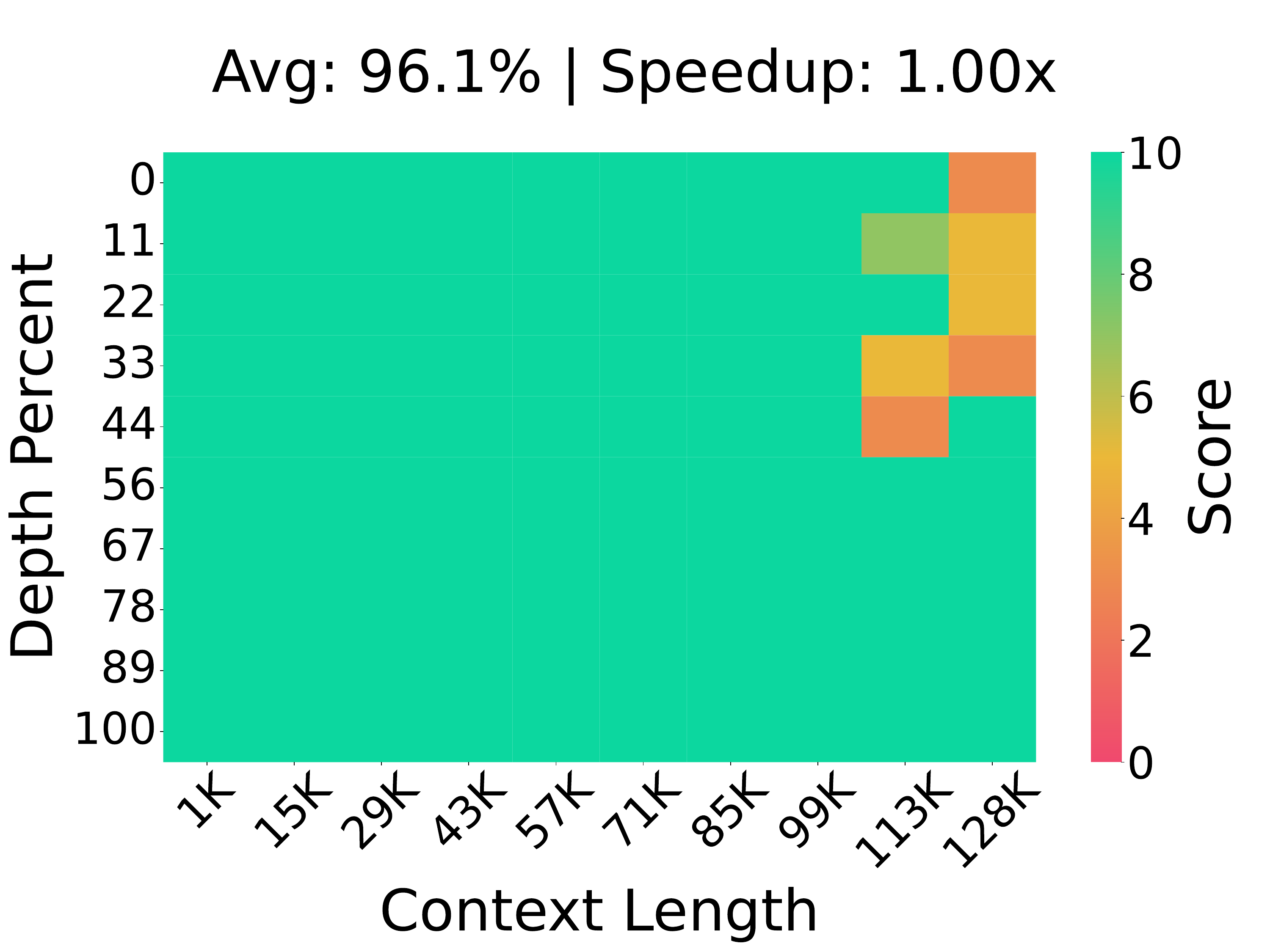}
}
}
\subfigure[MInference.]{
\scalebox{0.3}{
\includegraphics[width=\linewidth]{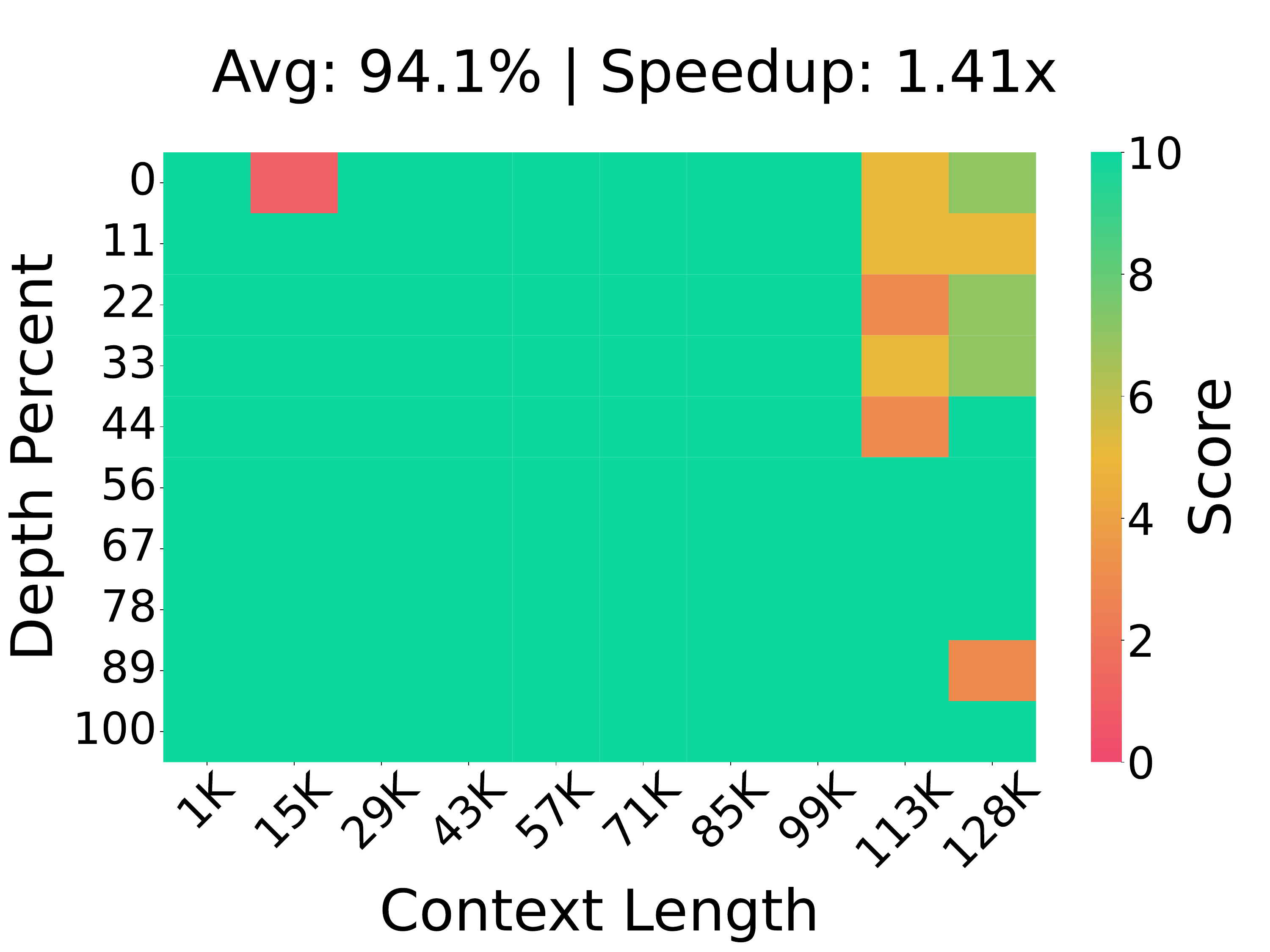}
}
}
\subfigure[FlexPrefill.]{
\scalebox{0.3}{
\includegraphics[width=\linewidth]{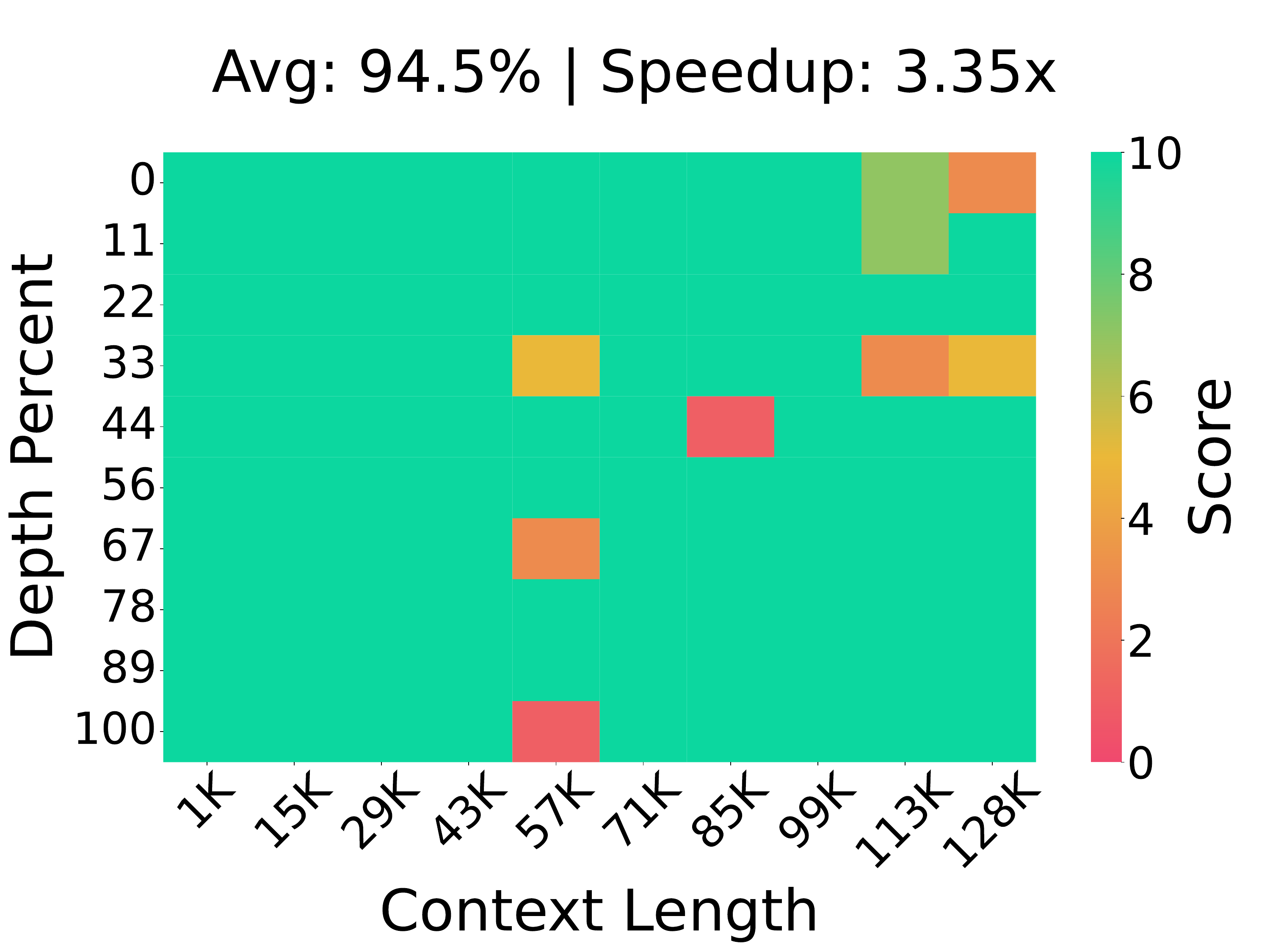}
}
}
\subfigure[SpargeAttn.]{
\scalebox{0.3}{
\includegraphics[width=\linewidth]{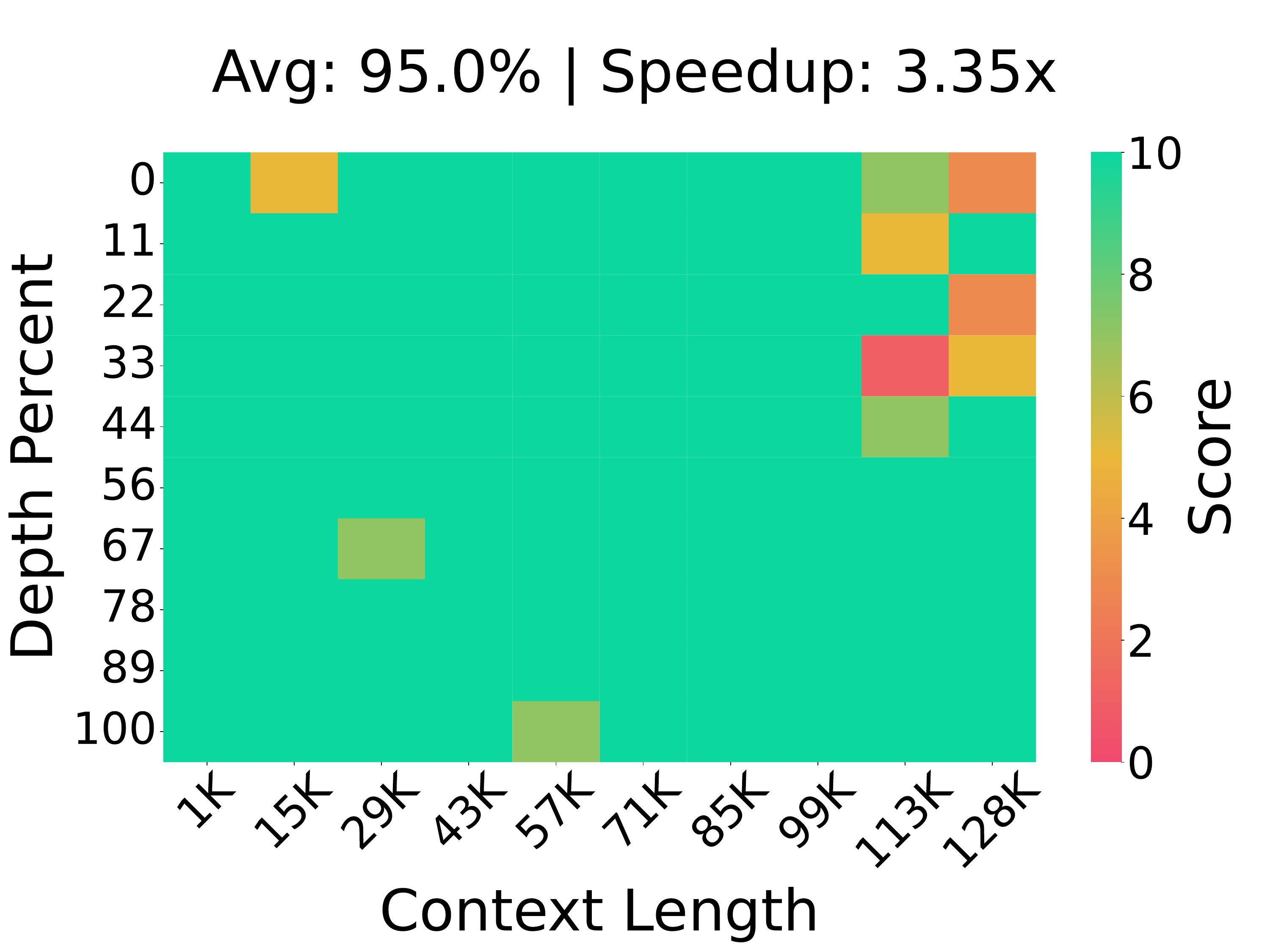}
}
}
\subfigure[\ourMethod.]{
\scalebox{0.3}{
\includegraphics[width=\linewidth]{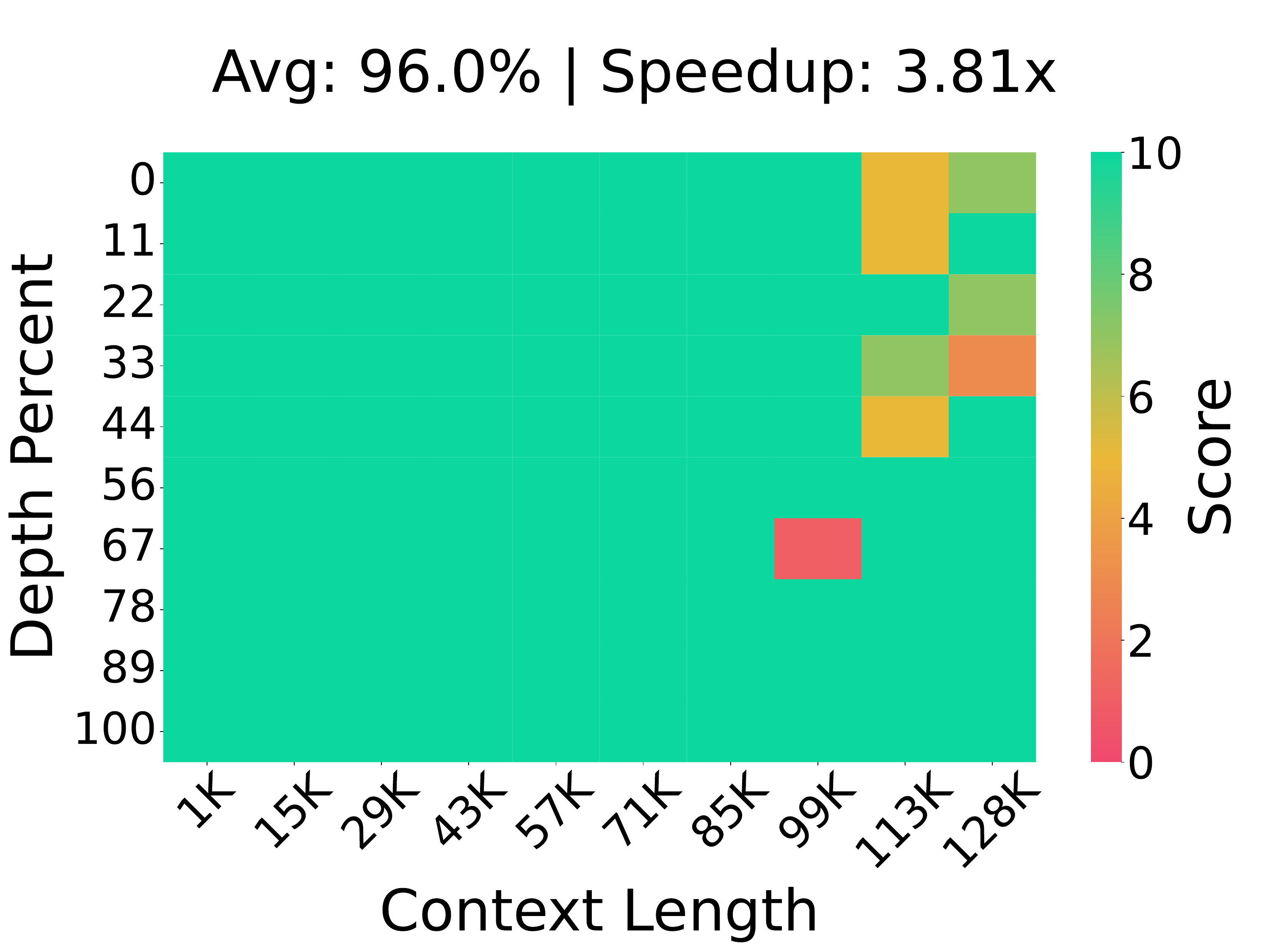}
}
}
\caption{Needle-In-A-Haystack evaluation results.}
\label{fig:niah}
\end{figure}

\paragraph{Needle-In-A-Haystack}
We evaluate the Needle-In-A-Haystack~(NIAH) task using Llama-3.1, with results visualized in~\Cref{fig:niah}. 
The average score and end-to-end speedup for each method are annotated above their respective plots.
Our method achieves a 3.81$\times$ speedup with only a $0.1\%$ drop in average score compared to FlashAttention2, outperforming all other sparse attention baselines.

\begin{figure}[b]
\vspace{-10pt}
\centering
\subfigure[Single input speedup.]{
\label{fig:ablationAndSpeedup:speedup}
\scalebox{0.54}{
\includegraphics[width=\linewidth]{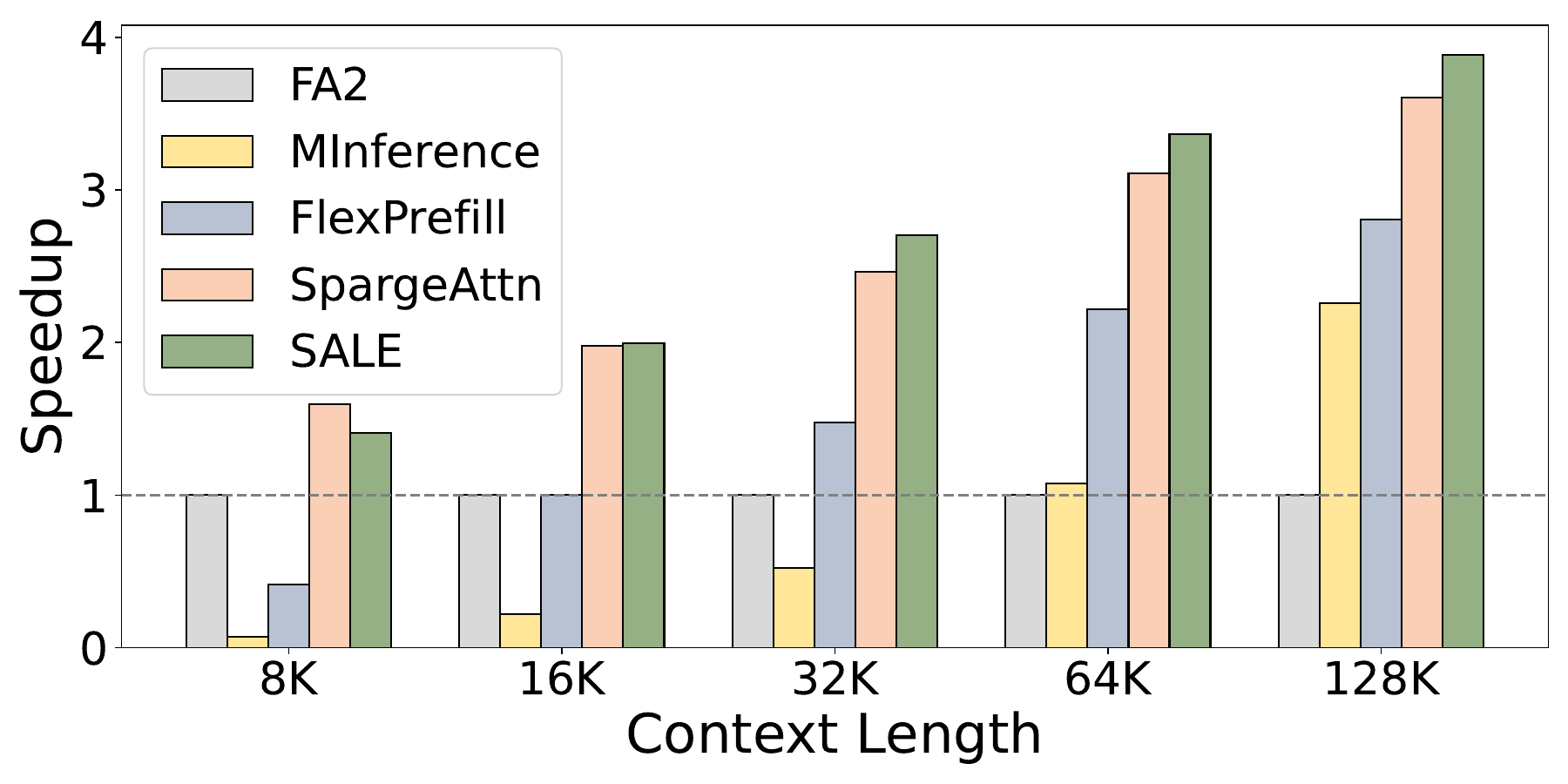}
}
}
\subfigure[Effect of calibration.]{
\label{fig:ablation:calib}
\scalebox{0.36}{
\includegraphics[width=\linewidth]{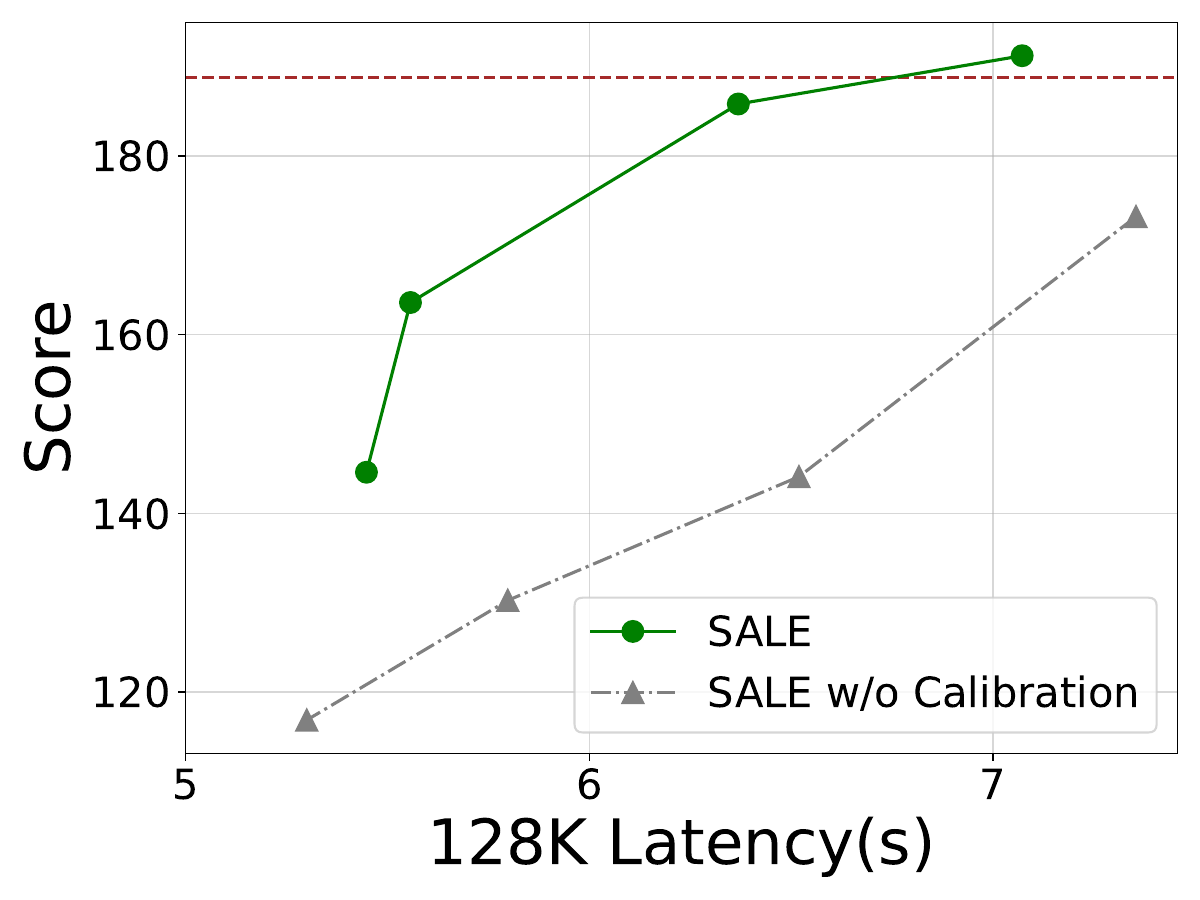}
}
}
\caption{(a) Speedup in single-input processing. (b) Comparison between SALE v.s. SALE w/o Calibration on InfiniteBench. The brown horizontal dashed line represents the score achieved by FlashAttention2. }
\label{fig:ablationAndSpeedup}
\end{figure}
\subsection{Efficiency evaluation}
\paragraph{Single input speedup}
We first compare the latency of different methods when processing a single input. The results are presented in~\Cref{fig:ablationAndSpeedup:speedup}. We conduct experiments using Llama-3.1 and report the speedup of each method relative to FlashAttention2. To illustrate how latency scales with the number of tokens, we prepare five input samples of different lengths. These samples are obtained by truncating a single 128K-length input from the Needle-In-A-Haystack task.

\begin{figure}[t]
\centering
\subfigure[Trade-off on LongBench.]{
\label{fig:tradeoff:Long}
\scalebox{0.30}{
\includegraphics[width=\linewidth]{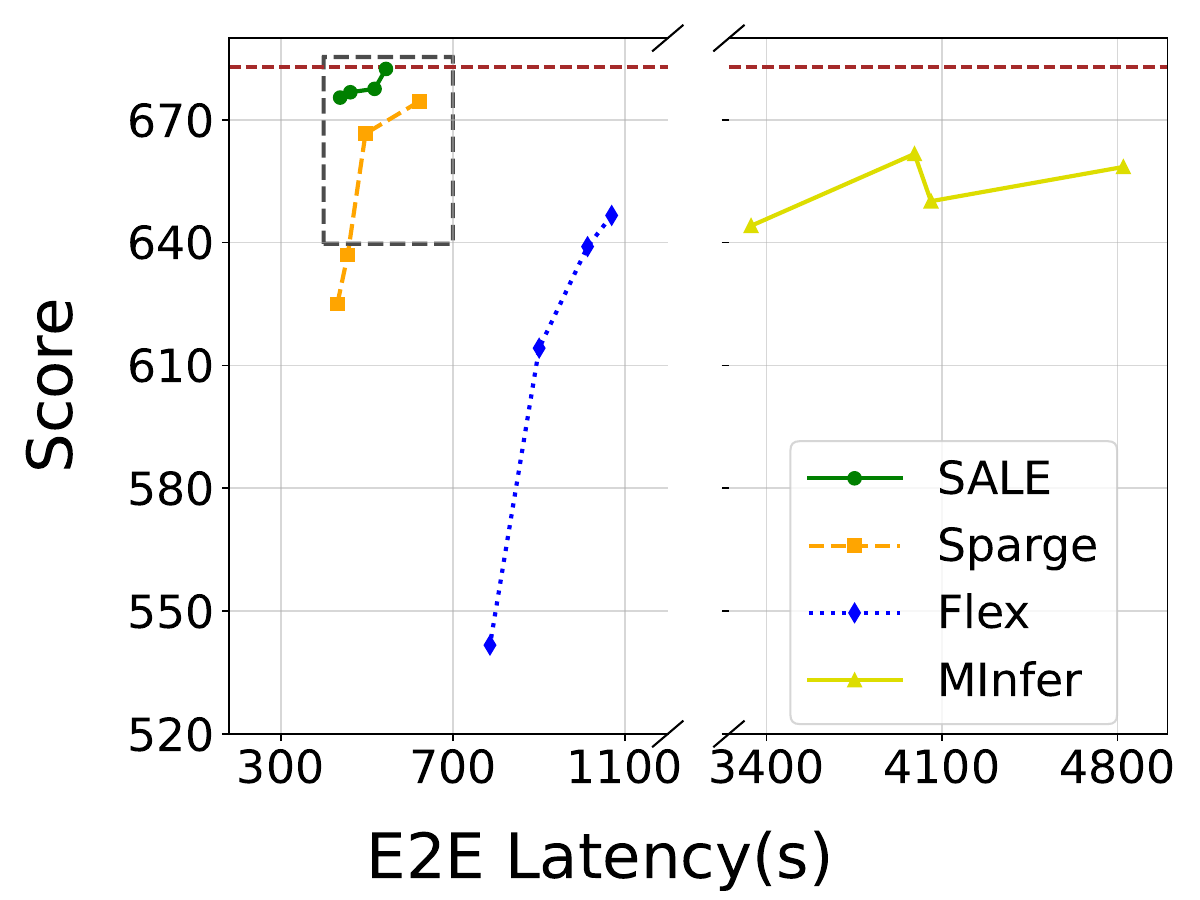}
}
}
\subfigure[Zoom in.]{
\label{fig:tradeoff:LongDetail}
\scalebox{0.225}{
\includegraphics[width=\linewidth]{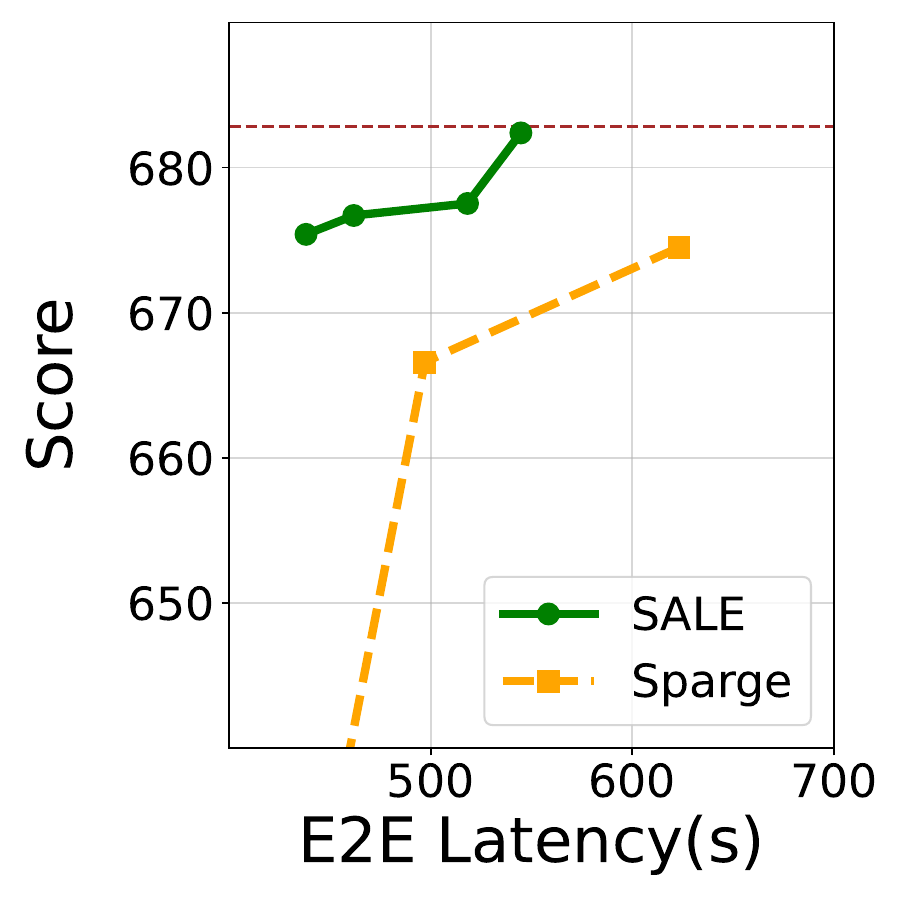}
}
}
\subfigure[Trade-off on InfiniteBench.]{
\label{fig:tradeoff:Infi}
\scalebox{0.375}{
\includegraphics[width=\linewidth]{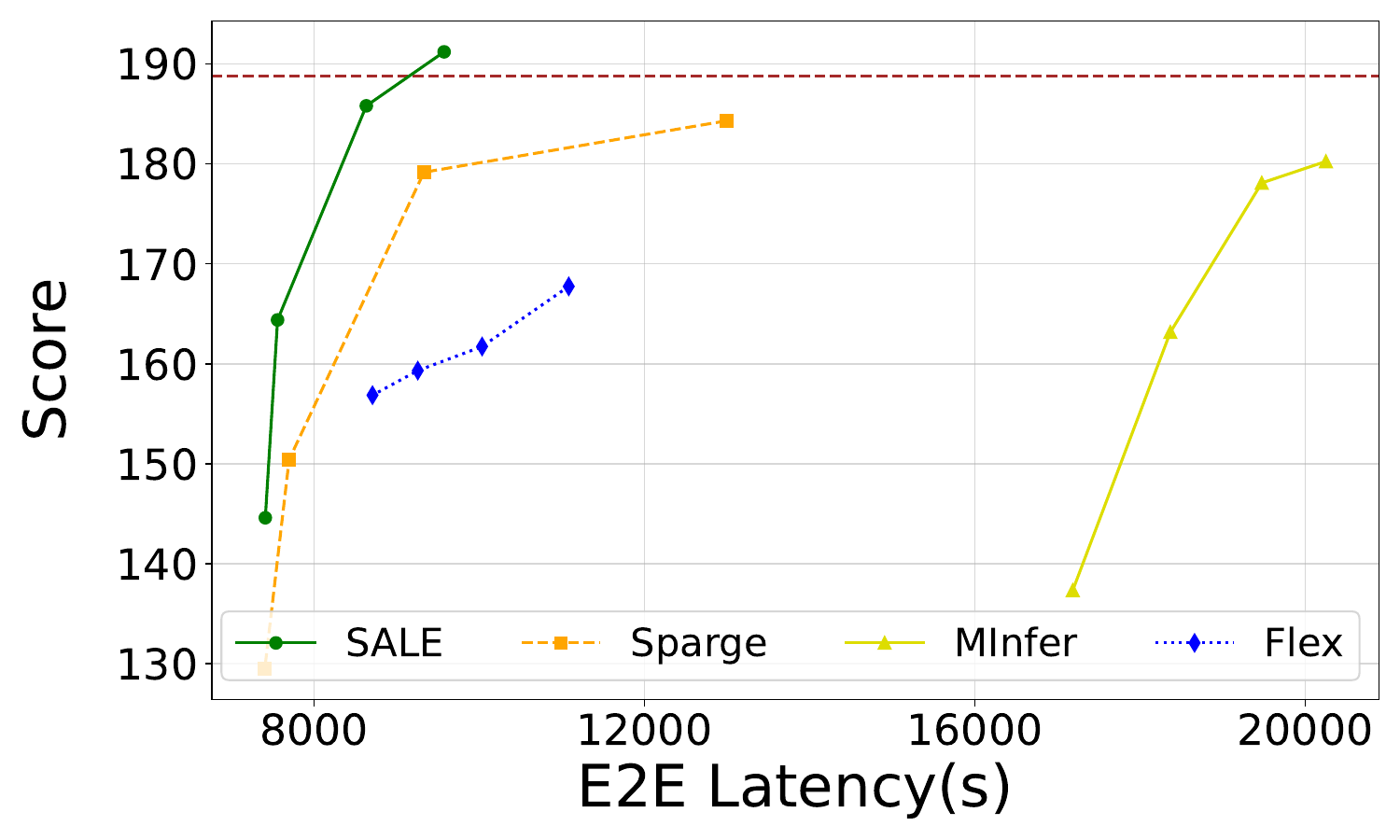}
}
}
\caption{
Evaluation of accuracy-efficiency trade-offs. The brown horizontal dashed line represents the score achieved by FlashAttention2. (a) Performance on LongBench under different sparsity levels. (b) A magnified view focusing on the region enclosed by the dashed box in (a). (c) Performance on InfiniteBench under different sparsity levels.
}
\label{fig:tradeoff1}
\vspace{-10pt}
\end{figure}

Our method demonstrates consistent speedups over FlashAttention2 across all sequence lengths while outperforming all sparse attention baselines in most cases.
Notably, \ourMethod exhibits greater speedup as context length increases, benefiting from sparser attention patterns.
\paragraph{Accuracy vs efficiency} 
We adjust the computation budget of each method following the approach described in~\Cref{sec:baselines} to analyze the accuracy-efficiency trade-offs. 
Considering that the speedup achieved by dynamic sparse attention methods may vary depending on the input content, we evaluate the end-to-end latency of all methods on both LongBench and InfiniteBench for comprehensive comparison.
The results, shown in~\Cref{fig:tradeoff1}, demonstrate the superior performance of our method on both datasets.
\subsection{Ablation study}
In this section, we evaluate the latency of each stage in \ourMethod and assess the impact of per-head threshold calibration. Additional analysis results are provided in the appendix.

\paragraph{Latency breakdown}
We report the latency breakdown results of \ourMethod under various input lengths in~\Cref{tab:latencyBreakdown}. 
All experiments use Llama-3.1, with reported timings reflecting end-to-end execution across all 32 model layers.
In the second-to-last line, we show the execution time ratio of Quantization and \lowBitPass operations relative to full attention latency.
In the final line, we present the speedup of \fullPass compared to full attention.
The results demonstrate that our method introduces acceptable computational overhead, with its relative cost decreasing as sequence length grows.
Furthermore, \fullPass shows greater speedups with longer context lengths, reflecting improved sparsity level at scale.
\begin{table}[htbp]
\vspace{-10pt}
\centering
\scriptsize
\setlength{\abovecaptionskip}{8pt}
\setlength{\tabcolsep}{0pt}
\setlength{\extrarowheight}{1pt}
\captionof{table}{Latency breakdown (ms).}
\label{tab:latencyBreakdown}
\begin{tabular}{p{3.5cm}<{\centering}|p{1cm}<{\centering}p{1cm}<{\centering}p{1cm}<{\centering}p{1cm}<{\centering}p{1cm}<{\centering}}
\toprule
\textbf{Context length} & \textbf{8K} & \textbf{16K} & \textbf{32K} & \textbf{64K} & \textbf{128K} \\
\midrule 
Quantization   &11     &21     &47    &99    &208   \\
\lowBitPass    &14     &48     &166   &634   &2562  \\
\fullPass      &51     &137    &378   &1117  &3599  \\
FA2            &106    &416    &1597  &6224  &24731 \\
\midrule
Overhead ratio &23.9$\%$ &16.7$\%$ &13.3$\%$ &11.5$\%$ &11.1$\%$  \\
\fullPass speedup &2.08$\times$ &3.04$\times$ &4.23$\times$ &5.57$\times$ & 6.87$\times$ \\
\bottomrule
\end{tabular}
\vspace{-10pt}
\end{table}

\paragraph{Threshold calibration}
To demonstrate the performance gain brought by per-head threshold calibration, we set all heads in Llama-3.1 to share the same $\tau$, which is referred to as \textit{\ourMethod w/o Calibration}. 
As shown in~\Cref{fig:ablation:calib}, per-head threshold calibration yields substantial performance gains.
\section{Conclusion}

In this paper, we propose a block-\textbf{S}parse \textbf{A}ttention technique based on \textbf{L}ow-bit \textbf{E}stimation. By performing fine-grained estimation of the attention map, we achieve a better accuracy-efficiency trade-off. Specifically, we estimate the attention weights using low-bit quantized queries and keys, and assess the importance of query-key pairs using our \textit{Relative Attention Score} metric. Furthermore, we introduce several CUDA kernel optimization techniques to ensure the efficiency of sparse mask construction on hardware. These components allow our method to efficiently and accurately analyze attention patterns. Experimental results demonstrate that our approach achieves the best trade-off among existing sparse attention baselines, delivering a speedup of at least 3.36$\times$ when processing sequences longer than 64K tokens while maintaining negligible accuracy loss.
{
\small
\bibliographystyle{ieeetr}
\bibliography{reference}    

\begin{thebibliography}{10}

\bibitem{kryściński2022booksumcollectiondatasetslongform}
W.~Kryściński, N.~Rajani, D.~Agarwal, C.~Xiong, and D.~Radev, ``Booksum: A collection of datasets for long-form narrative summarization,'' 2022.

\bibitem{porwal2023transformer}
S.~Porwal, L.~Bewoor, and V.~Deshpande, ``Transformer based implementation for automatic book summarization,'' {\em arXiv preprint arXiv:2301.07057}, 2023.

\bibitem{chang2024booookscore}
Y.~Chang, K.~Lo, T.~Goyal, and M.~Iyyer, ``Booookscore: A systematic exploration of book-length summarization in the era of {LLM}s,'' in {\em The Twelfth International Conference on Learning Representations}, 2024.

\bibitem{caciularu-etal-2023-peek}
A.~Caciularu, M.~Peters, J.~Goldberger, I.~Dagan, and A.~Cohan, ``Peek across: Improving multi-document modeling via cross-document question-answering,'' in {\em Proceedings of the 61st Annual Meeting of the Association for Computational Linguistics (Volume 1: Long Papers)} (A.~Rogers, J.~Boyd-Graber, and N.~Okazaki, eds.), (Toronto, Canada), pp.~1970--1989, Association for Computational Linguistics, July 2023.

\bibitem{pang2022qualityquestionansweringlong}
R.~Y. Pang, A.~Parrish, N.~Joshi, N.~Nangia, J.~Phang, A.~Chen, V.~Padmakumar, J.~Ma, J.~Thompson, H.~He, and S.~R. Bowman, ``Quality: Question answering with long input texts, yes!,'' 2022.

\bibitem{fan2019eli5longformquestion}
A.~Fan, Y.~Jernite, E.~Perez, D.~Grangier, J.~Weston, and M.~Auli, ``Eli5: Long form question answering,'' 2019.

\bibitem{wang2024teachingcodellmsuse}
C.~Wang, J.~Zhang, Y.~Feng, T.~Li, W.~Sun, Y.~Liu, and X.~Peng, ``Teaching code llms to use autocompletion tools in repository-level code generation,'' 2024.

\bibitem{wang2024rlcoderreinforcementlearningrepositorylevel}
Y.~Wang, Y.~Wang, D.~Guo, J.~Chen, R.~Zhang, Y.~Ma, and Z.~Zheng, ``Rlcoder: Reinforcement learning for repository-level code completion,'' 2024.

\bibitem{grattafiori2024llama3herdmodels}
A.~Grattafiori, A.~Dubey, A.~Jauhri, A.~Pandey, A.~Kadian, A.~Al-Dahle, A.~Letman, A.~Mathur, A.~Schelten, A.~Vaughan, A.~Yang, A.~Fan, A.~Goyal, A.~Hartshorn, A.~Yang, A.~Mitra, A.~Sravankumar, A.~Korenev, A.~Hinsvark, A.~Rao, A.~Zhang, A.~Rodriguez, A.~Gregerson, A.~Spataru, B.~Roziere, B.~Biron, B.~Tang, B.~Chern, C.~Caucheteux, C.~Nayak, C.~Bi, C.~Marra, C.~McConnell, C.~Keller, C.~Touret, C.~Wu, C.~Wong, C.~C. Ferrer, C.~Nikolaidis, D.~Allonsius, D.~Song, D.~Pintz, D.~Livshits, D.~Wyatt, D.~Esiobu, D.~Choudhary, D.~Mahajan, D.~Garcia-Olano, D.~Perino, D.~Hupkes, E.~Lakomkin, E.~AlBadawy, E.~Lobanova, E.~Dinan, E.~M. Smith, F.~Radenovic, F.~Guzmán, F.~Zhang, G.~Synnaeve, G.~Lee, G.~L. Anderson, G.~Thattai, G.~Nail, G.~Mialon, G.~Pang, G.~Cucurell, H.~Nguyen, H.~Korevaar, H.~Xu, H.~Touvron, I.~Zarov, I.~A. Ibarra, I.~Kloumann, I.~Misra, I.~Evtimov, J.~Zhang, J.~Copet, J.~Lee, J.~Geffert, J.~Vranes, J.~Park, J.~Mahadeokar, J.~Shah, J.~van~der Linde, J.~Billock, J.~Hong, J.~Lee, J.~Fu, J.~Chi, J.~Huang,
  J.~Liu, J.~Wang, J.~Yu, J.~Bitton, J.~Spisak, J.~Park, J.~Rocca, J.~Johnstun, J.~Saxe, J.~Jia, K.~V. Alwala, K.~Prasad, K.~Upasani, K.~Plawiak, K.~Li, K.~Heafield, K.~Stone, K.~El-Arini, K.~Iyer, K.~Malik, K.~Chiu, K.~Bhalla, K.~Lakhotia, L.~Rantala-Yeary, L.~van~der Maaten, L.~Chen, L.~Tan, L.~Jenkins, L.~Martin, L.~Madaan, L.~Malo, L.~Blecher, L.~Landzaat, L.~de~Oliveira, M.~Muzzi, M.~Pasupuleti, M.~Singh, M.~Paluri, M.~Kardas, M.~Tsimpoukelli, M.~Oldham, M.~Rita, M.~Pavlova, M.~Kambadur, M.~Lewis, M.~Si, M.~K. Singh, M.~Hassan, N.~Goyal, N.~Torabi, N.~Bashlykov, N.~Bogoychev, N.~Chatterji, N.~Zhang, O.~Duchenne, O.~Çelebi, P.~Alrassy, P.~Zhang, P.~Li, P.~Vasic, P.~Weng, P.~Bhargava, P.~Dubal, P.~Krishnan, P.~S. Koura, P.~Xu, Q.~He, Q.~Dong, R.~Srinivasan, R.~Ganapathy, R.~Calderer, R.~S. Cabral, R.~Stojnic, R.~Raileanu, R.~Maheswari, R.~Girdhar, R.~Patel, R.~Sauvestre, R.~Polidoro, R.~Sumbaly, R.~Taylor, R.~Silva, R.~Hou, R.~Wang, S.~Hosseini, S.~Chennabasappa, S.~Singh, S.~Bell, S.~S. Kim, S.~Edunov,
  S.~Nie, S.~Narang, S.~Raparthy, S.~Shen, S.~Wan, S.~Bhosale, S.~Zhang, S.~Vandenhende, S.~Batra, S.~Whitman, S.~Sootla, S.~Collot, S.~Gururangan, S.~Borodinsky, T.~Herman, T.~Fowler, T.~Sheasha, T.~Georgiou, T.~Scialom, T.~Speckbacher, T.~Mihaylov, T.~Xiao, U.~Karn, V.~Goswami, V.~Gupta, V.~Ramanathan, V.~Kerkez, V.~Gonguet, V.~Do, V.~Vogeti, V.~Albiero, V.~Petrovic, W.~Chu, W.~Xiong, W.~Fu, W.~Meers, X.~Martinet, X.~Wang, X.~Wang, X.~E. Tan, X.~Xia, X.~Xie, X.~Jia, X.~Wang, Y.~Goldschlag, Y.~Gaur, Y.~Babaei, Y.~Wen, Y.~Song, Y.~Zhang, Y.~Li, Y.~Mao, Z.~D. Coudert, Z.~Yan, Z.~Chen, Z.~Papakipos, A.~Singh, A.~Srivastava, A.~Jain, A.~Kelsey, A.~Shajnfeld, A.~Gangidi, A.~Victoria, A.~Goldstand, A.~Menon, A.~Sharma, A.~Boesenberg, A.~Baevski, A.~Feinstein, A.~Kallet, A.~Sangani, A.~Teo, A.~Yunus, A.~Lupu, A.~Alvarado, A.~Caples, A.~Gu, A.~Ho, A.~Poulton, A.~Ryan, A.~Ramchandani, A.~Dong, A.~Franco, A.~Goyal, A.~Saraf, A.~Chowdhury, A.~Gabriel, A.~Bharambe, A.~Eisenman, A.~Yazdan, B.~James, B.~Maurer,
  B.~Leonhardi, B.~Huang, B.~Loyd, B.~D. Paola, B.~Paranjape, B.~Liu, B.~Wu, B.~Ni, B.~Hancock, B.~Wasti, B.~Spence, B.~Stojkovic, B.~Gamido, B.~Montalvo, C.~Parker, C.~Burton, C.~Mejia, C.~Liu, C.~Wang, C.~Kim, C.~Zhou, C.~Hu, C.-H. Chu, C.~Cai, C.~Tindal, C.~Feichtenhofer, C.~Gao, D.~Civin, D.~Beaty, D.~Kreymer, D.~Li, D.~Adkins, D.~Xu, D.~Testuggine, D.~David, D.~Parikh, D.~Liskovich, D.~Foss, D.~Wang, D.~Le, D.~Holland, E.~Dowling, E.~Jamil, E.~Montgomery, E.~Presani, E.~Hahn, E.~Wood, E.-T. Le, E.~Brinkman, E.~Arcaute, E.~Dunbar, E.~Smothers, F.~Sun, F.~Kreuk, F.~Tian, F.~Kokkinos, F.~Ozgenel, F.~Caggioni, F.~Kanayet, F.~Seide, G.~M. Florez, G.~Schwarz, G.~Badeer, G.~Swee, G.~Halpern, G.~Herman, G.~Sizov, Guangyi, Zhang, G.~Lakshminarayanan, H.~Inan, H.~Shojanazeri, H.~Zou, H.~Wang, H.~Zha, H.~Habeeb, H.~Rudolph, H.~Suk, H.~Aspegren, H.~Goldman, H.~Zhan, I.~Damlaj, I.~Molybog, I.~Tufanov, I.~Leontiadis, I.-E. Veliche, I.~Gat, J.~Weissman, J.~Geboski, J.~Kohli, J.~Lam, J.~Asher, J.-B. Gaya, J.~Marcus,
  J.~Tang, J.~Chan, J.~Zhen, J.~Reizenstein, J.~Teboul, J.~Zhong, J.~Jin, J.~Yang, J.~Cummings, J.~Carvill, J.~Shepard, J.~McPhie, J.~Torres, J.~Ginsburg, J.~Wang, K.~Wu, K.~H. U, K.~Saxena, K.~Khandelwal, K.~Zand, K.~Matosich, K.~Veeraraghavan, K.~Michelena, K.~Li, K.~Jagadeesh, K.~Huang, K.~Chawla, K.~Huang, L.~Chen, L.~Garg, L.~A, L.~Silva, L.~Bell, L.~Zhang, L.~Guo, L.~Yu, L.~Moshkovich, L.~Wehrstedt, M.~Khabsa, M.~Avalani, M.~Bhatt, M.~Mankus, M.~Hasson, M.~Lennie, M.~Reso, M.~Groshev, M.~Naumov, M.~Lathi, M.~Keneally, M.~Liu, M.~L. Seltzer, M.~Valko, M.~Restrepo, M.~Patel, M.~Vyatskov, M.~Samvelyan, M.~Clark, M.~Macey, M.~Wang, M.~J. Hermoso, M.~Metanat, M.~Rastegari, M.~Bansal, N.~Santhanam, N.~Parks, N.~White, N.~Bawa, N.~Singhal, N.~Egebo, N.~Usunier, N.~Mehta, N.~P. Laptev, N.~Dong, N.~Cheng, O.~Chernoguz, O.~Hart, O.~Salpekar, O.~Kalinli, P.~Kent, P.~Parekh, P.~Saab, P.~Balaji, P.~Rittner, P.~Bontrager, P.~Roux, P.~Dollar, P.~Zvyagina, P.~Ratanchandani, P.~Yuvraj, Q.~Liang, R.~Alao, R.~Rodriguez,
  R.~Ayub, R.~Murthy, R.~Nayani, R.~Mitra, R.~Parthasarathy, R.~Li, R.~Hogan, R.~Battey, R.~Wang, R.~Howes, R.~Rinott, S.~Mehta, S.~Siby, S.~J. Bondu, S.~Datta, S.~Chugh, S.~Hunt, S.~Dhillon, S.~Sidorov, S.~Pan, S.~Mahajan, S.~Verma, S.~Yamamoto, S.~Ramaswamy, S.~Lindsay, S.~Lindsay, S.~Feng, S.~Lin, S.~C. Zha, S.~Patil, S.~Shankar, S.~Zhang, S.~Zhang, S.~Wang, S.~Agarwal, S.~Sajuyigbe, S.~Chintala, S.~Max, S.~Chen, S.~Kehoe, S.~Satterfield, S.~Govindaprasad, S.~Gupta, S.~Deng, S.~Cho, S.~Virk, S.~Subramanian, S.~Choudhury, S.~Goldman, T.~Remez, T.~Glaser, T.~Best, T.~Koehler, T.~Robinson, T.~Li, T.~Zhang, T.~Matthews, T.~Chou, T.~Shaked, V.~Vontimitta, V.~Ajayi, V.~Montanez, V.~Mohan, V.~S. Kumar, V.~Mangla, V.~Ionescu, V.~Poenaru, V.~T. Mihailescu, V.~Ivanov, W.~Li, W.~Wang, W.~Jiang, W.~Bouaziz, W.~Constable, X.~Tang, X.~Wu, X.~Wang, X.~Wu, X.~Gao, Y.~Kleinman, Y.~Chen, Y.~Hu, Y.~Jia, Y.~Qi, Y.~Li, Y.~Zhang, Y.~Zhang, Y.~Adi, Y.~Nam, Yu, Wang, Y.~Zhao, Y.~Hao, Y.~Qian, Y.~Li, Y.~He, Z.~Rait, Z.~DeVito,
  Z.~Rosnbrick, Z.~Wen, Z.~Yang, Z.~Zhao, and Z.~Ma, ``The llama 3 herd of models,'' 2024.

\bibitem{yang2025qwen2}
A.~Yang, B.~Yu, C.~Li, D.~Liu, F.~Huang, H.~Huang, J.~Jiang, J.~Tu, J.~Zhang, J.~Zhou, {\em et~al.}, ``Qwen2. 5-1m technical report,'' {\em arXiv preprint arXiv:2501.15383}, 2025.

\bibitem{gemmateam2025gemma3technicalreport}
G.~Team, A.~Kamath, J.~Ferret, S.~Pathak, N.~Vieillard, R.~Merhej, S.~Perrin, T.~Matejovicova, A.~Ramé, M.~Rivière, L.~Rouillard, T.~Mesnard, G.~Cideron, J.~bastien Grill, S.~Ramos, E.~Yvinec, M.~Casbon, E.~Pot, I.~Penchev, G.~Liu, F.~Visin, K.~Kenealy, L.~Beyer, X.~Zhai, A.~Tsitsulin, R.~Busa-Fekete, A.~Feng, N.~Sachdeva, B.~Coleman, Y.~Gao, B.~Mustafa, I.~Barr, E.~Parisotto, D.~Tian, M.~Eyal, C.~Cherry, J.-T. Peter, D.~Sinopalnikov, S.~Bhupatiraju, R.~Agarwal, M.~Kazemi, D.~Malkin, R.~Kumar, D.~Vilar, I.~Brusilovsky, J.~Luo, A.~Steiner, A.~Friesen, A.~Sharma, A.~Sharma, A.~M. Gilady, A.~Goedeckemeyer, A.~Saade, A.~Feng, A.~Kolesnikov, A.~Bendebury, A.~Abdagic, A.~Vadi, A.~György, A.~S. Pinto, A.~Das, A.~Bapna, A.~Miech, A.~Yang, A.~Paterson, A.~Shenoy, A.~Chakrabarti, B.~Piot, B.~Wu, B.~Shahriari, B.~Petrini, C.~Chen, C.~L. Lan, C.~A. Choquette-Choo, C.~Carey, C.~Brick, D.~Deutsch, D.~Eisenbud, D.~Cattle, D.~Cheng, D.~Paparas, D.~S. Sreepathihalli, D.~Reid, D.~Tran, D.~Zelle, E.~Noland, E.~Huizenga,
  E.~Kharitonov, F.~Liu, G.~Amirkhanyan, G.~Cameron, H.~Hashemi, H.~Klimczak-Plucińska, H.~Singh, H.~Mehta, H.~T. Lehri, H.~Hazimeh, I.~Ballantyne, I.~Szpektor, I.~Nardini, J.~Pouget-Abadie, J.~Chan, J.~Stanton, J.~Wieting, J.~Lai, J.~Orbay, J.~Fernandez, J.~Newlan, J.~yeong Ji, J.~Singh, K.~Black, K.~Yu, K.~Hui, K.~Vodrahalli, K.~Greff, L.~Qiu, M.~Valentine, M.~Coelho, M.~Ritter, M.~Hoffman, M.~Watson, M.~Chaturvedi, M.~Moynihan, M.~Ma, N.~Babar, N.~Noy, N.~Byrd, N.~Roy, N.~Momchev, N.~Chauhan, N.~Sachdeva, O.~Bunyan, P.~Botarda, P.~Caron, P.~K. Rubenstein, P.~Culliton, P.~Schmid, P.~G. Sessa, P.~Xu, P.~Stanczyk, P.~Tafti, R.~Shivanna, R.~Wu, R.~Pan, R.~Rokni, R.~Willoughby, R.~Vallu, R.~Mullins, S.~Jerome, S.~Smoot, S.~Girgin, S.~Iqbal, S.~Reddy, S.~Sheth, S.~Põder, S.~Bhatnagar, S.~R. Panyam, S.~Eiger, S.~Zhang, T.~Liu, T.~Yacovone, T.~Liechty, U.~Kalra, U.~Evci, V.~Misra, V.~Roseberry, V.~Feinberg, V.~Kolesnikov, W.~Han, W.~Kwon, X.~Chen, Y.~Chow, Y.~Zhu, Z.~Wei, Z.~Egyed, V.~Cotruta, M.~Giang, P.~Kirk,
  A.~Rao, K.~Black, N.~Babar, J.~Lo, E.~Moreira, L.~G. Martins, O.~Sanseviero, L.~Gonzalez, Z.~Gleicher, T.~Warkentin, V.~Mirrokni, E.~Senter, E.~Collins, J.~Barral, Z.~Ghahramani, R.~Hadsell, Y.~Matias, D.~Sculley, S.~Petrov, N.~Fiedel, N.~Shazeer, O.~Vinyals, J.~Dean, D.~Hassabis, K.~Kavukcuoglu, C.~Farabet, E.~Buchatskaya, J.-B. Alayrac, R.~Anil, Dmitry, Lepikhin, S.~Borgeaud, O.~Bachem, A.~Joulin, A.~Andreev, C.~Hardin, R.~Dadashi, and L.~Hussenot, ``Gemma 3 technical report,'' 2025.

\bibitem{deepseekai2025deepseekv3technicalreport}
DeepSeek-AI, A.~Liu, B.~Feng, B.~Xue, B.~Wang, B.~Wu, C.~Lu, C.~Zhao, C.~Deng, C.~Zhang, C.~Ruan, D.~Dai, D.~Guo, D.~Yang, D.~Chen, D.~Ji, E.~Li, F.~Lin, F.~Dai, F.~Luo, G.~Hao, G.~Chen, G.~Li, H.~Zhang, H.~Bao, H.~Xu, H.~Wang, H.~Zhang, H.~Ding, H.~Xin, H.~Gao, H.~Li, H.~Qu, J.~L. Cai, J.~Liang, J.~Guo, J.~Ni, J.~Li, J.~Wang, J.~Chen, J.~Chen, J.~Yuan, J.~Qiu, J.~Li, J.~Song, K.~Dong, K.~Hu, K.~Gao, K.~Guan, K.~Huang, K.~Yu, L.~Wang, L.~Zhang, L.~Xu, L.~Xia, L.~Zhao, L.~Wang, L.~Zhang, M.~Li, M.~Wang, M.~Zhang, M.~Zhang, M.~Tang, M.~Li, N.~Tian, P.~Huang, P.~Wang, P.~Zhang, Q.~Wang, Q.~Zhu, Q.~Chen, Q.~Du, R.~J. Chen, R.~L. Jin, R.~Ge, R.~Zhang, R.~Pan, R.~Wang, R.~Xu, R.~Zhang, R.~Chen, S.~S. Li, S.~Lu, S.~Zhou, S.~Chen, S.~Wu, S.~Ye, S.~Ye, S.~Ma, S.~Wang, S.~Zhou, S.~Yu, S.~Zhou, S.~Pan, T.~Wang, T.~Yun, T.~Pei, T.~Sun, W.~L. Xiao, W.~Zeng, W.~Zhao, W.~An, W.~Liu, W.~Liang, W.~Gao, W.~Yu, W.~Zhang, X.~Q. Li, X.~Jin, X.~Wang, X.~Bi, X.~Liu, X.~Wang, X.~Shen, X.~Chen, X.~Zhang, X.~Chen, X.~Nie, X.~Sun,
  X.~Wang, X.~Cheng, X.~Liu, X.~Xie, X.~Liu, X.~Yu, X.~Song, X.~Shan, X.~Zhou, X.~Yang, X.~Li, X.~Su, X.~Lin, Y.~K. Li, Y.~Q. Wang, Y.~X. Wei, Y.~X. Zhu, Y.~Zhang, Y.~Xu, Y.~Xu, Y.~Huang, Y.~Li, Y.~Zhao, Y.~Sun, Y.~Li, Y.~Wang, Y.~Yu, Y.~Zheng, Y.~Zhang, Y.~Shi, Y.~Xiong, Y.~He, Y.~Tang, Y.~Piao, Y.~Wang, Y.~Tan, Y.~Ma, Y.~Liu, Y.~Guo, Y.~Wu, Y.~Ou, Y.~Zhu, Y.~Wang, Y.~Gong, Y.~Zou, Y.~He, Y.~Zha, Y.~Xiong, Y.~Ma, Y.~Yan, Y.~Luo, Y.~You, Y.~Liu, Y.~Zhou, Z.~F. Wu, Z.~Z. Ren, Z.~Ren, Z.~Sha, Z.~Fu, Z.~Xu, Z.~Huang, Z.~Zhang, Z.~Xie, Z.~Zhang, Z.~Hao, Z.~Gou, Z.~Ma, Z.~Yan, Z.~Shao, Z.~Xu, Z.~Wu, Z.~Zhang, Z.~Li, Z.~Gu, Z.~Zhu, Z.~Liu, Z.~Li, Z.~Xie, Z.~Song, Z.~Gao, and Z.~Pan, ``Deepseek-v3 technical report,'' 2025.

\bibitem{vaswani2017attention}
A.~Vaswani, N.~Shazeer, N.~Parmar, J.~Uszkoreit, L.~Jones, A.~N. Gomez, {\L}.~Kaiser, and I.~Polosukhin, ``Attention is all you need,'' {\em Advances in neural information processing systems}, vol.~30, 2017.

\bibitem{fu2024challengesdeployinglongcontexttransformers}
Y.~Fu, ``Challenges in deploying long-context transformers: A theoretical peak performance analysis,'' 2024.

\bibitem{jiang2024minference}
H.~Jiang, Y.~Li, C.~Zhang, Q.~Wu, X.~Luo, S.~Ahn, Z.~Han, A.~Abdi, D.~Li, C.-Y. Lin, {\em et~al.}, ``Minference 1.0: Accelerating pre-filling for long-context llms via dynamic sparse attention,'' {\em Advances in Neural Information Processing Systems}, vol.~37, pp.~52481--52515, 2024.

\bibitem{DBLP:journals/corr/abs-2404-02690}
Y.~Deng, Z.~Song, and C.~Yang, ``Attention is naturally sparse with gaussian distributed input,'' {\em CoRR}, vol.~abs/2404.02690, 2024.

\bibitem{child2019generating}
R.~Child, S.~Gray, A.~Radford, and I.~Sutskever, ``Generating long sequences with sparse transformers,'' {\em arXiv preprint arXiv:1904.10509}, 2019.

\bibitem{zaheer2020big}
M.~Zaheer, G.~Guruganesh, K.~A. Dubey, J.~Ainslie, C.~Alberti, S.~Ontanon, P.~Pham, A.~Ravula, Q.~Wang, L.~Yang, {\em et~al.}, ``Big bird: Transformers for longer sequences,'' {\em Advances in neural information processing systems}, vol.~33, pp.~17283--17297, 2020.

\bibitem{beltagy2020longformer}
I.~Beltagy, M.~E. Peters, and A.~Cohan, ``Longformer: The long-document transformer,'' {\em arXiv preprint arXiv:2004.05150}, 2020.

\bibitem{xiao2024efficient}
G.~Xiao, Y.~Tian, B.~Chen, S.~Han, and M.~Lewis, ``Efficient streaming language models with attention sinks,'' in {\em The Twelfth International Conference on Learning Representations}, 2024.

\bibitem{han2023lm}
C.~Han, Q.~Wang, H.~Peng, W.~Xiong, Y.~Chen, H.~Ji, and S.~Wang, ``Lm-infinite: Zero-shot extreme length generalization for large language models,'' {\em arXiv preprint arXiv:2308.16137}, 2023.

\bibitem{zhu2024sampleattention}
Q.~Zhu, J.~Duan, C.~Chen, S.~Liu, X.~Li, G.~Feng, X.~Lv, H.~Cao, X.~Chuanfu, X.~Zhang, {\em et~al.}, ``Sampleattention: Near-lossless acceleration of long context llm inference with adaptive structured sparse attention,'' {\em arXiv preprint arXiv:2406.15486}, 2024.

\bibitem{lai2025flexprefill}
X.~Lai, J.~Lu, Y.~Luo, Y.~Ma, and X.~Zhou, ``Flexprefill: A context-aware sparse attention mechanism for efficient long-sequence inference,'' in {\em The Thirteenth International Conference on Learning Representations}, 2025.

\bibitem{zhang2025spargeattn}
J.~Zhang, C.~Xiang, H.~Huang, J.~Wei, H.~Xi, J.~Zhu, and J.~Chen, ``Spargeattn: Accurate sparse attention accelerating any model inference,'' {\em arXiv preprint arXiv:2502.18137}, 2025.

\bibitem{lee2025a}
H.~Lee, G.~Park, Y.~Lee, J.~Suh, J.~Kim, W.~Jeong, B.~Kim, H.~Lee, M.~Jeon, and S.~J. Hwang, ``A training-free sub-quadratic cost transformer model serving framework with hierarchically pruned attention,'' in {\em The Thirteenth International Conference on Learning Representations}, 2025.

\bibitem{gu2025when}
X.~Gu, T.~Pang, C.~Du, Q.~Liu, F.~Zhang, C.~Du, Y.~Wang, and M.~Lin, ``When attention sink emerges in language models: An empirical view,'' in {\em The Thirteenth International Conference on Learning Representations}, 2025.

\bibitem{zhang2023h2o}
Z.~Zhang, Y.~Sheng, T.~Zhou, T.~Chen, L.~Zheng, R.~Cai, Z.~Song, Y.~Tian, C.~R{\'e}, C.~Barrett, {\em et~al.}, ``H2o: Heavy-hitter oracle for efficient generative inference of large language models,'' {\em Advances in Neural Information Processing Systems}, vol.~36, pp.~34661--34710, 2023.

\bibitem{li2024snapkv}
Y.~Li, Y.~Huang, B.~Yang, B.~Venkitesh, A.~Locatelli, H.~Ye, T.~Cai, P.~Lewis, and D.~Chen, ``Snapkv: Llm knows what you are looking for before generation,'' {\em Advances in Neural Information Processing Systems}, vol.~37, pp.~22947--22970, 2024.

\bibitem{zhang2024pqcache}
H.~Zhang, X.~Ji, Y.~Chen, F.~Fu, X.~Miao, X.~Nie, W.~Chen, and B.~Cui, ``Pqcache: Product quantization-based kvcache for long context llm inference,'' {\em arXiv preprint arXiv:2407.12820}, 2024.

\bibitem{liu2025speculative}
J.~Liu, B.~Chen, and C.~Zhang, ``Speculative prefill: Turbocharging ttft with lightweight and training-free token importance estimation,'' {\em arXiv preprint arXiv:2502.02789}, 2025.

\bibitem{yang2024qwen2}
A.~Yang, B.~Yang, B.~Zhang, B.~Hui, B.~Zheng, B.~Yu, C.~Li, D.~Liu, F.~Huang, H.~Wei, {\em et~al.}, ``Qwen2. 5 technical report,'' {\em arXiv preprint arXiv:2412.15115}, 2024.

\bibitem{jha2024characterizing}
S.~Jha, L.~E. Erdogan, S.~Kim, K.~Keutzer, and A.~Gholami, ``Characterizing prompt compression methods for long context inference,'' {\em arXiv preprint arXiv:2407.08892}, 2024.

\bibitem{shi2024discovering}
Z.~Shi, Y.~Ming, X.-P. Nguyen, Y.~Liang, and S.~Joty, ``Discovering the gems in early layers: Accelerating long-context llms with 1000x input token reduction,'' {\em arXiv preprint arXiv:2409.17422}, 2024.

\bibitem{jiang2023llmlingua}
H.~Jiang, Q.~Wu, C.-Y. Lin, Y.~Yang, and L.~Qiu, ``Llmlingua: Compressing prompts for accelerated inference of large language models,'' {\em arXiv preprint arXiv:2310.05736}, 2023.

\bibitem{li2023compressing}
Y.~Li, B.~Dong, C.~Lin, and F.~Guerin, ``Compressing context to enhance inference efficiency of large language models,'' {\em arXiv preprint arXiv:2310.06201}, 2023.

\bibitem{yuan2024kv}
J.~Yuan, H.~Liu, S.~Zhong, Y.-N. Chuang, S.~Li, G.~Wang, D.~Le, H.~Jin, V.~Chaudhary, Z.~Xu, {\em et~al.}, ``Kv cache compression, but what must we give in return? a comprehensive benchmark of long context capable approaches,'' {\em arXiv preprint arXiv:2407.01527}, 2024.

\bibitem{fu2024moa}
T.~Fu, H.~Huang, X.~Ning, G.~Zhang, B.~Chen, T.~Wu, H.~Wang, Z.~Huang, S.~Li, S.~Yan, {\em et~al.}, ``Moa: Mixture of sparse attention for automatic large language model compression,'' {\em arXiv preprint arXiv:2406.14909}, 2024.

\bibitem{xiao2024duoattention}
G.~Xiao, J.~Tang, J.~Zuo, J.~Guo, S.~Yang, H.~Tang, Y.~Fu, and S.~Han, ``Duoattention: Efficient long-context llm inference with retrieval and streaming heads,'' {\em arXiv preprint arXiv:2410.10819}, 2024.

\bibitem{gao2024seerattention}
Y.~Gao, Z.~Zeng, D.~Du, S.~Cao, P.~Zhou, J.~Qi, J.~Lai, H.~K.-H. So, T.~Cao, F.~Yang, {\em et~al.}, ``Seerattention: Learning intrinsic sparse attention in your llms,'' {\em arXiv preprint arXiv:2410.13276}, 2024.

\bibitem{yuan2025native}
J.~Yuan, H.~Gao, D.~Dai, J.~Luo, L.~Zhao, Z.~Zhang, Z.~Xie, Y.~Wei, L.~Wang, Z.~Xiao, {\em et~al.}, ``Native sparse attention: Hardware-aligned and natively trainable sparse attention,'' {\em arXiv preprint arXiv:2502.11089}, 2025.

\bibitem{lu2025moba}
E.~Lu, Z.~Jiang, J.~Liu, Y.~Du, T.~Jiang, C.~Hong, S.~Liu, W.~He, E.~Yuan, Y.~Wang, {\em et~al.}, ``Moba: Mixture of block attention for long-context llms,'' {\em arXiv preprint arXiv:2502.13189}, 2025.

\bibitem{peng2023rwkv}
B.~Peng, E.~Alcaide, Q.~Anthony, A.~Albalak, S.~Arcadinho, S.~Biderman, H.~Cao, X.~Cheng, M.~Chung, M.~Grella, {\em et~al.}, ``Rwkv: Reinventing rnns for the transformer era,'' {\em arXiv preprint arXiv:2305.13048}, 2023.

\bibitem{yang2023gated}
S.~Yang, B.~Wang, Y.~Shen, R.~Panda, and Y.~Kim, ``Gated linear attention transformers with hardware-efficient training,'' {\em arXiv preprint arXiv:2312.06635}, 2023.

\bibitem{gu2024mamba}
A.~Gu and T.~Dao, ``Mamba: Linear-time sequence modeling with selective state spaces,'' in {\em First Conference on Language Modeling}, 2024.

\bibitem{10.5555/3692070.3692469}
T.~Dao and A.~Gu, ``Transformers are ssms: generalized models and efficient algorithms through structured state space duality,'' in {\em Proceedings of the 41st International Conference on Machine Learning}, ICML'24, JMLR.org, 2024.

\bibitem{ribar2023sparq}
L.~Ribar, I.~Chelombiev, L.~Hudlass-Galley, C.~Blake, C.~Luschi, and D.~Orr, ``Sparq attention: Bandwidth-efficient llm inference,'' {\em arXiv preprint arXiv:2312.04985}, 2023.

\bibitem{lee2024infinigen}
W.~Lee, J.~Lee, J.~Seo, and J.~Sim, ``$\{$InfiniGen$\}$: Efficient generative inference of large language models with dynamic $\{$KV$\}$ cache management,'' in {\em 18th USENIX Symposium on Operating Systems Design and Implementation (OSDI 24)}, pp.~155--172, 2024.

\bibitem{chen2025magicpig}
Z.~Chen, R.~Sadhukhan, Z.~Ye, Y.~Zhou, J.~Zhang, N.~Nolte, Y.~Tian, M.~Douze, L.~Bottou, Z.~Jia, and B.~Chen, ``Magic{PIG}: {LSH} sampling for efficient {LLM} generation,'' in {\em The Thirteenth International Conference on Learning Representations}, 2025.

\bibitem{liu2024retrievalattention}
D.~Liu, M.~Chen, B.~Lu, H.~Jiang, Z.~Han, Q.~Zhang, Q.~Chen, C.~Zhang, B.~Ding, K.~Zhang, {\em et~al.}, ``Retrievalattention: Accelerating long-context llm inference via vector retrieval,'' {\em arXiv preprint arXiv:2409.10516}, 2024.

\bibitem{liu2023scissorhands}
Z.~Liu, A.~Desai, F.~Liao, W.~Wang, V.~Xie, Z.~Xu, A.~Kyrillidis, and A.~Shrivastava, ``Scissorhands: Exploiting the persistence of importance hypothesis for llm kv cache compression at test time,'' {\em Advances in Neural Information Processing Systems}, vol.~36, pp.~52342--52364, 2023.

\bibitem{ge2023model}
S.~Ge, Y.~Zhang, L.~Liu, M.~Zhang, J.~Han, and J.~Gao, ``Model tells you what to discard: Adaptive kv cache compression for llms,'' {\em arXiv preprint arXiv:2310.01801}, 2023.

\bibitem{devoto2024simple}
A.~Devoto, Y.~Zhao, S.~Scardapane, and P.~Minervini, ``A simple and effective $ l\_2 $ norm-based strategy for kv cache compression,'' {\em arXiv preprint arXiv:2406.11430}, 2024.

\bibitem{zhang2024cam}
Y.~Zhang, Y.~Du, G.~Luo, Y.~Zhong, Z.~Zhang, S.~Liu, and R.~Ji, ``Cam: Cache merging for memory-efficient llms inference,'' in {\em Forty-first International Conference on Machine Learning}, 2024.

\bibitem{zandieh2024subgen}
A.~Zandieh, I.~Han, V.~Mirrokni, and A.~Karbasi, ``Subgen: Token generation in sublinear time and memory,'' {\em arXiv preprint arXiv:2402.06082}, 2024.

\bibitem{dao2022flashattention}
T.~Dao, D.~Fu, S.~Ermon, A.~Rudra, and C.~R{\'e}, ``Flashattention: Fast and memory-efficient exact attention with io-awareness,'' {\em Advances in neural information processing systems}, vol.~35, pp.~16344--16359, 2022.

\bibitem{dao2023flashattention}
T.~Dao, ``Flashattention-2: Faster attention with better parallelism and work partitioning,'' {\em arXiv preprint arXiv:2307.08691}, 2023.

\bibitem{shah2024flashattention}
J.~Shah, G.~Bikshandi, Y.~Zhang, V.~Thakkar, P.~Ramani, and T.~Dao, ``Flashattention-3: Fast and accurate attention with asynchrony and low-precision,'' {\em Advances in Neural Information Processing Systems}, vol.~37, pp.~68658--68685, 2024.

\bibitem{sanovar2024lean}
R.~Sanovar, S.~Bharadwaj, R.~S. Amant, V.~R{\"u}hle, and S.~Rajmohan, ``Lean attention: Hardware-aware scalable attention mechanism for the decode-phase of transformers,'' {\em arXiv preprint arXiv:2405.10480}, 2024.

\bibitem{zhang2024sageattention2}
J.~Zhang, H.~Huang, P.~Zhang, J.~Wei, J.~Zhu, and J.~Chen, ``Sageattention2 technical report: Accurate 4 bit attention for plug-and-play inference acceleration,'' {\em arXiv preprint arXiv:2411.10958}, 2024.

\bibitem{zhang2025sageattention}
J.~Zhang, J.~wei, P.~Zhang, J.~Zhu, and J.~Chen, ``Sageattention: Accurate 8-bit attention for plug-and-play inference acceleration,'' in {\em The Thirteenth International Conference on Learning Representations}, 2025.

\bibitem{10.1145/3315508.3329973}
P.~Tillet, H.~T. Kung, and D.~Cox, ``Triton: an intermediate language and compiler for tiled neural network computations,'' in {\em Proceedings of the 3rd ACM SIGPLAN International Workshop on Machine Learning and Programming Languages}, MAPL 2019, (New York, NY, USA), p.~10–19, Association for Computing Machinery, 2019.

\bibitem{wolf-etal-2020-transformers}
T.~Wolf, L.~Debut, V.~Sanh, J.~Chaumond, C.~Delangue, A.~Moi, P.~Cistac, T.~Rault, R.~Louf, M.~Funtowicz, J.~Davison, S.~Shleifer, P.~von Platen, C.~Ma, Y.~Jernite, J.~Plu, C.~Xu, T.~L. Scao, S.~Gugger, M.~Drame, Q.~Lhoest, and A.~M. Rush, ``Transformers: State-of-the-art natural language processing,'' in {\em Proceedings of the 2020 Conference on Empirical Methods in Natural Language Processing: System Demonstrations}, (Online), pp.~38--45, Association for Computational Linguistics, Oct. 2020.

\bibitem{shoeybi2019megatron}
M.~Shoeybi, M.~Patwary, R.~Puri, P.~LeGresley, J.~Casper, and B.~Catanzaro, ``Megatron-lm: Training multi-billion parameter language models using model parallelism,'' {\em arXiv preprint arXiv:1909.08053}, 2019.

\bibitem{li2023distflashattn}
D.~Li, R.~Shao, A.~Xie, E.~P. Xing, X.~Ma, I.~Stoica, J.~E. Gonzalez, and H.~Zhang, ``Distflashattn: Distributed memory-efficient attention for long-context llms training,'' {\em arXiv preprint arXiv:2310.03294}, 2023.

\bibitem{bai-etal-2024-longbench}
Y.~Bai, X.~Lv, J.~Zhang, H.~Lyu, J.~Tang, Z.~Huang, Z.~Du, X.~Liu, A.~Zeng, L.~Hou, Y.~Dong, J.~Tang, and J.~Li, ``{L}ong{B}ench: A bilingual, multitask benchmark for long context understanding,'' in {\em Proceedings of the 62nd Annual Meeting of the Association for Computational Linguistics (Volume 1: Long Papers)} (L.-W. Ku, A.~Martins, and V.~Srikumar, eds.), (Bangkok, Thailand), pp.~3119--3137, Association for Computational Linguistics, Aug. 2024.

\bibitem{zhang2024inftybenchextendinglongcontext}
X.~Zhang, Y.~Chen, S.~Hu, Z.~Xu, J.~Chen, M.~K. Hao, X.~Han, Z.~L. Thai, S.~Wang, Z.~Liu, and M.~Sun, ``$\infty$bench: Extending long context evaluation beyond 100k tokens,'' 2024.

\bibitem{Greg2024niah}
G.~Kamradt, ``Needle in a haystack - pressure testing llms,'' 2023.

\end{thebibliography}
}
\appendix
\newpage
\appendix

\section{Limitation}
Due to our method's reliance on high-throughput 4-bit Tensor Core instructions to accelerate the \lowBitPass, it may lose its performance advantage on hardware that does not support efficient 4-bit matrix multiplication. 

Moreover, our current implementation is limited to approximating attention weights using Int4 quantization.
Additional adaptations would be needed to deploy our method on hardware that supports FP4 GEMM or LUT-based low-bit GEMM.
We leave it as our future work.

\section{Broader impact}
\ourMethod significantly reduces the computational cost of the long-context LLM \prefill, thereby lowering deployment costs and enabling broader adoption of AI technologies. 
This advancement also facilitates the development of applications that rely on processing long contexts.
Additionally, it contributes to a reduction in energy consumption of LLM services.

\section{Additional implementation details}
We select five input samples from the Retrieve.KV task in InfiniteBench to perform calibration for \ourMethod, and the final configuration must satisfy the error bound requirement across all five samples.
The per-head threshold calibration for Llama-3.1 on RTX4090 server takes approximately five minutes to complete.

For the local area discussed in~\Cref{sec:algo_design}, we set its size to be no smaller than 128 tokens.
Since the comparison results of \textit{Relative Attention Score} and $\tau$ for each thread are not visible to others, an all-reduce operation must be performed across all threads within the GPU thread block to aggregate these results, which incurs considerable overhead. 
To reduce the frequency of all-reduce operations, we group every four consecutive key blocks into a \textbf{segment} and perform result aggregation at the segment level. 
At the end of each row, any remaining key blocks~(less than 4 blocks) are also treated as local area blocks for implementation simplicity. 
As a result, the number of tokens in local area may exceed 128, but will not surpass 256. 

\section{Additional experiment details}
We use the same input samples to search the optimal hyperparameters for SpargeAttn, and use the first input sample to search sparse pattern configuration for MInference based on its open-source implementation.

During evaluation process, to ensure proper model behavior, we truncate samples that exceed the maximum context window length. 
Following common practice, we retain the tokens from both the beginning and the end of the sequence and remove those from the middle portion.

For the data format during model inference, we employed BFloat16 for FlexPrefill due to requirements specified in its repository, while Float16 was used for all other methods.

\begin{figure}[htbp]
\centering
\subfigure[Single input speedup]{
\label{fig:fp16_approx:speedup}
\scalebox{0.4}{
\includegraphics[width=\linewidth]{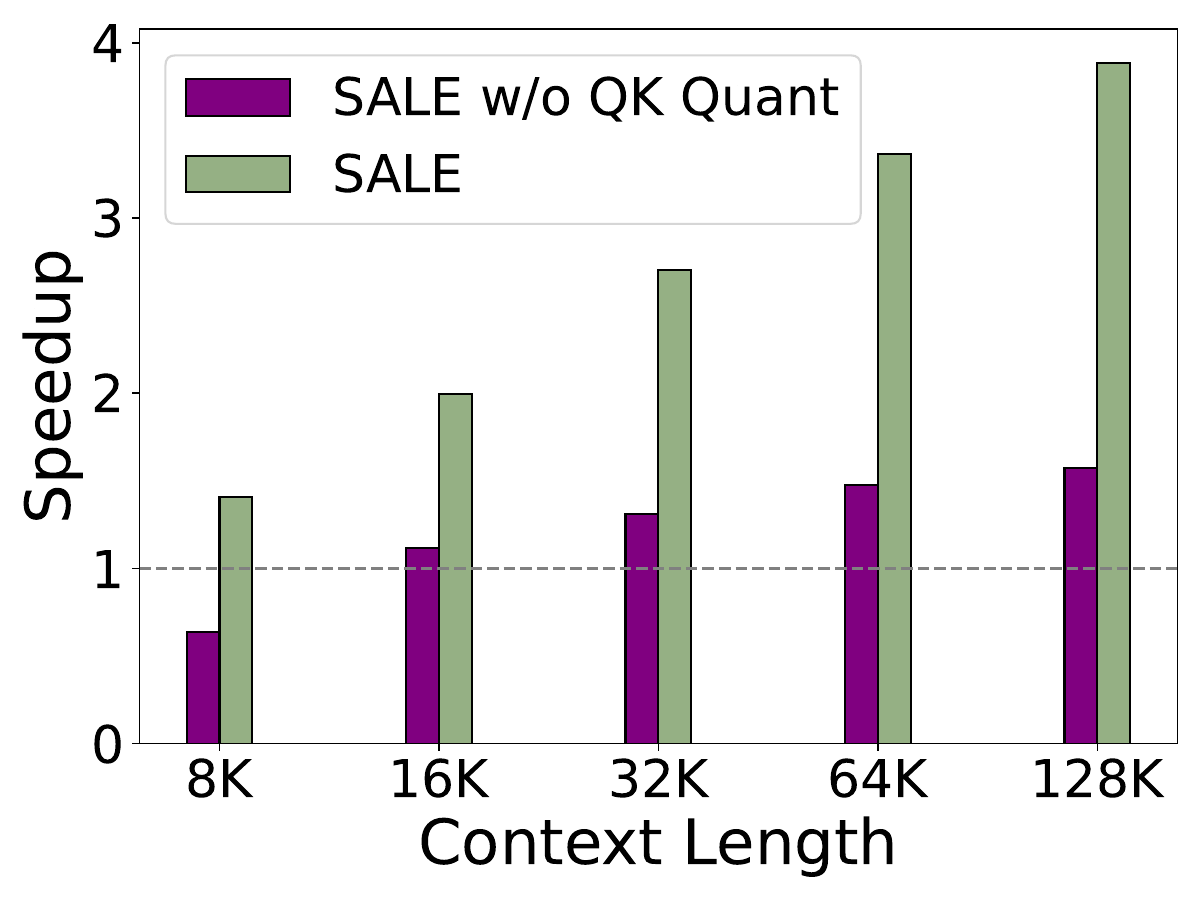}
}
}
\subfigure[Accuracy and sparsity result]{
\label{fig:fp16_approx:tradeoff}
\scalebox{0.4}{
\includegraphics[width=\linewidth]{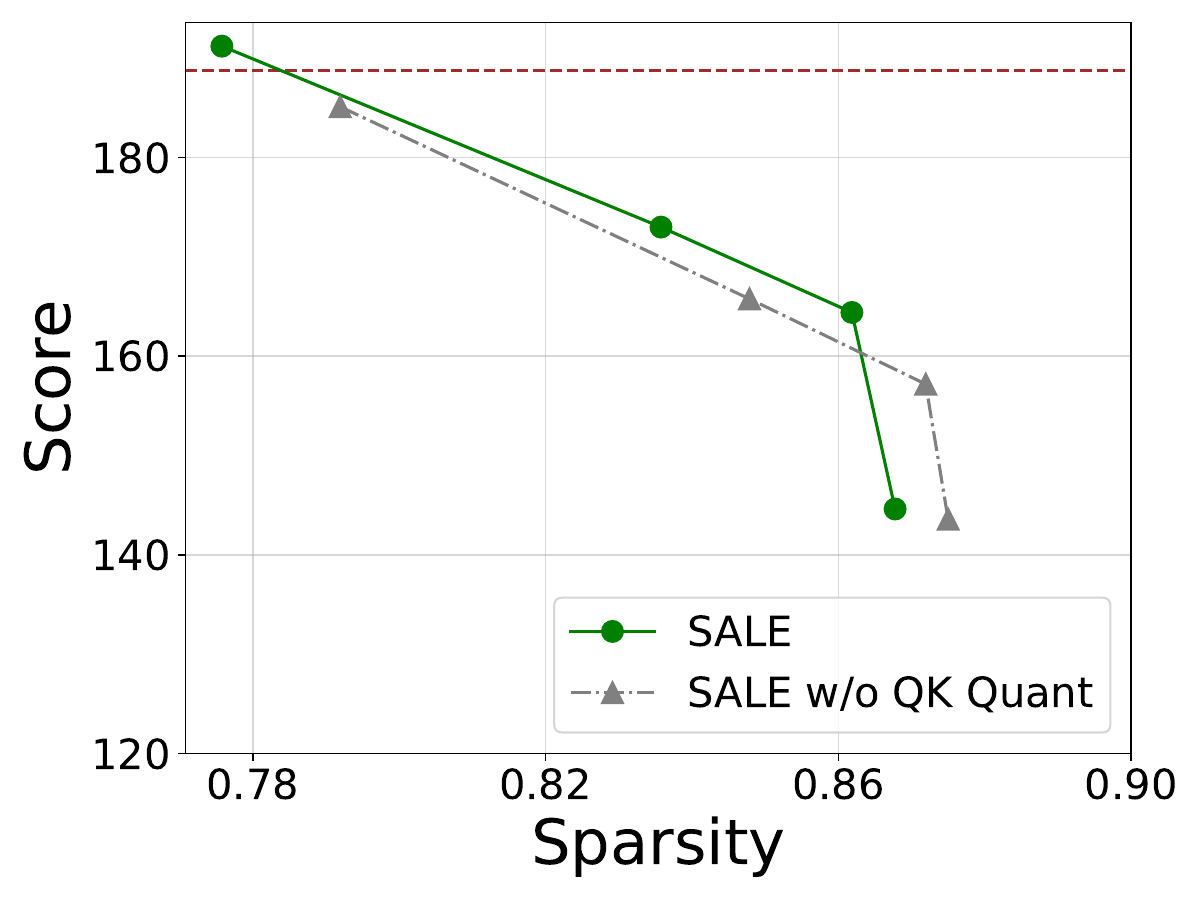}
}
}
\caption{
Comparison between \ourMethod and \ourMethod w/o QK Quant. (a)Single input speedup. (b) Comparison between \ourMethod v.s. \ourMethod w/o QK Quant on InfiniteBench. The brown horizontal dashed line represents the score achieved by FlashAttention2.
}
\label{fig:fp16_approx}
\vspace{-10pt}
\end{figure}

\section{Additional ablation studies}
To evaluate the effectiveness of 4-bit attention weight approximation, we further conducted experiments using original-precision~(16-bit) QK matrices to inspect the attention map, which is referred to as \textit{\ourMethod w/o QK Quant}.
The result is shown in ~\Cref{fig:fp16_approx}.
We measure the single input speedup of two methods under varying input lengths, using the same set of input samples as in the~\Cref{fig:ablationAndSpeedup:speedup}. The result indicates that using original-precision QK to estimate attention weights leads to a significant increase in computational overhead.

We further evaluate the accuracy and attention sparsity of both methods based on Llama-3.1, where corresponding data points for the two methods are obtained using the same $\theta$. 
We use the scores from InfiniteBench to represent accuracy.
Attention sparsity metric is defined as the ratio of the number of skipped attention blocks to the total number of attention blocks, and the results presented here are measured when processing contexts of 128K length. 
As observed, under identical hyperparameter settings, \textit{\ourMethod w/o QK Quant} achieves higher attention sparsity while showing a slight performance drop on InfiniteBench.
This may be attributed to the limited precision of current Int4 quantization techniques, which can cause certain approximated attention weights to exceed their true values, thereby leading to more blocks being selected.

\newpage

\end{document}